\documentclass[runningheads]{llncs}

\usepackage{eccv}

\usepackage{multirow}
\usepackage{colortbl}
\definecolor{Gray}{gray}{0.9}

\usepackage{xspace}
\usepackage{xfrac}
\usepackage{graphicx}
\usepackage{stfloats}
\usepackage{wrapfig}

\newcommand{\ours}{{HYPE}\xspace}
\newcommand{\ourss}{$\text{HYPE}_{\text{score}}$\xspace}

\newcommand{\vv}{\mathbf{v}}
\newcommand{\vx}{\mathbf{x}}
\newcommand{\vy}{\mathbf{y}}

\newcommand{\vO}{\mathbf{O}}
\newcommand{\bbR}{\mathbb{R}}

\newcommand{\ent}{$\epsilon_{*}$\xspace}
\newcommand{\enti}{$\epsilon_{i}$\xspace}
\newcommand{\entt}{$\epsilon_{t}$\xspace}
\newcommand{\negl}{$-d_{\mathcal{L}}$\xspace}
\newcommand{\cossim}{$cos(\theta)$\xspace}

\usepackage{eccvabbrv}

\usepackage{graphicx}
\usepackage{booktabs}

\usepackage[accsupp]{axessibility}  %
\usepackage{adjustbox}

\usepackage[pagebackref,breaklinks,colorlinks,citecolor=eccvblue]{hyperref}

\usepackage[capitalize,noabbrev]{cleveref}
\crefname{equation}{Eqn.}{Eqns.}
\creflabelformat{equation}{#2\textup{#1}#3}

\usepackage{orcidlink}

\usepackage[sort,numbers]{natbib}

\begin{document}
\abovedisplayskip=0.5em
\abovedisplayshortskip=0.5em
\belowdisplayskip=0.5em
\belowdisplayshortskip=0.5em

\title{HYPE: Hyperbolic Entailment Filtering for Underspecified Images and Texts}

\titlerunning{Hyperbolic Entailment Filtering for Underspecified Images and Texts}

\author{Wonjae Kim ~ Sanghyuk Chun ~ Taekyung Kim ~ Dongyoon Han ~ Sangdoo Yun}

\authorrunning{W. Kim et al.}

\institute{NAVER AI Lab}

\maketitle

\begin{abstract}
In an era where the volume of data drives the effectiveness of self-supervised learning, the specificity and clarity of data semantics play a crucial role in model training.
Addressing this, we introduce HYPerbolic Entailment filtering (HYPE), a novel methodology designed to meticulously extract modality-wise meaningful and well-aligned data from extensive, noisy image-text pair datasets.
Our approach leverages hyperbolic embeddings and the concept of entailment cones to evaluate and filter out samples with meaningless or underspecified semantics, focusing on enhancing the specificity of each data sample.
HYPE not only demonstrates a significant improvement in filtering efficiency but also sets a new state-of-the-art in the DataComp benchmark when combined with existing filtering techniques.
This breakthrough showcases the potential of HYPE to refine the data selection process, thereby contributing to the development of more accurate and efficient self-supervised learning models.
Additionally, the image specificity $\epsilon_{i}$ can be independently applied to induce an image-only dataset from an image-text or image-only data pool for training image-only self-supervised models and showed superior performance when compared to the dataset induced by CLIP score.
\end{abstract}    
\section{Introduction}
\label{sec:intro}

\begin{figure}[t]
    \centering
    \includegraphics[width=\linewidth]{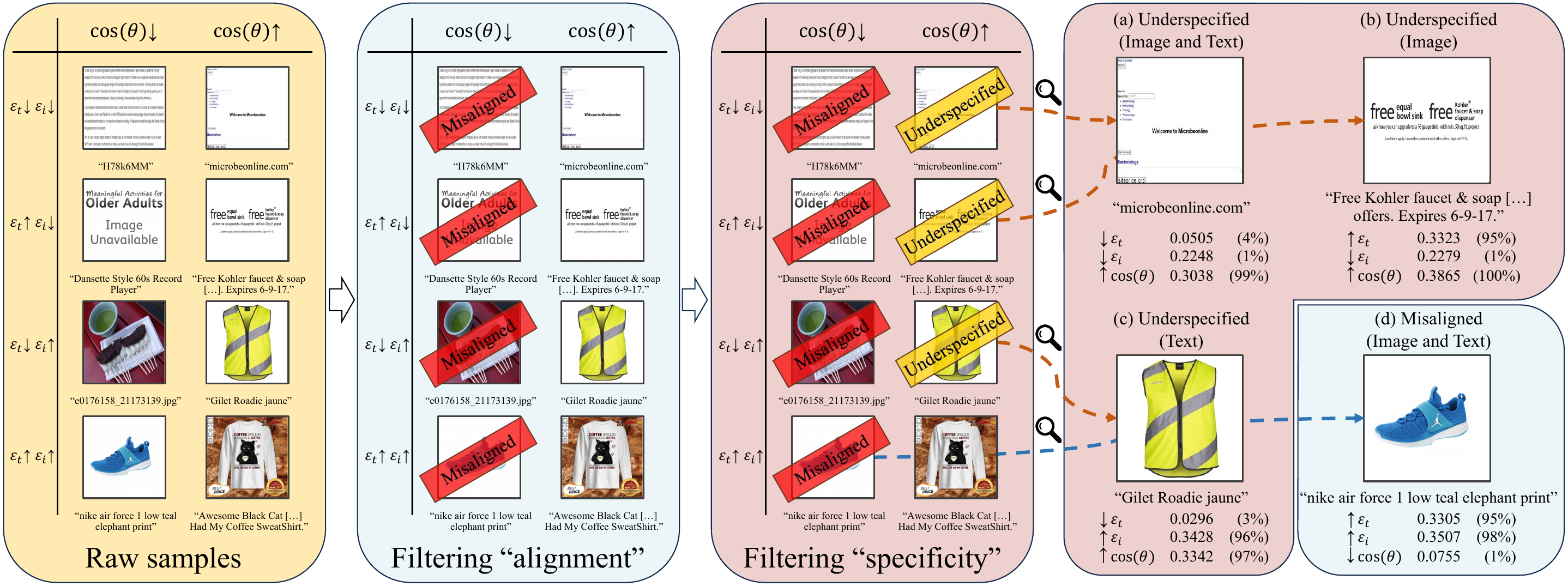}%
    \caption{Example of \ours filtering on the Datacomp small pool \cite{datacomp}.
    \ours leverages both uni-modal specificity (text specificity \entt and image specificity \enti) and cross-modal similarity (CLIP similarity \cossim as in this figure or negative Lorentzian distance \negl can be used instead) to effectively identify and eliminate misalignment and underspecification issues on noisy image-text pairs.
    Figures (a-c) show instances where image-text pairs exhibit high alignment yet are flagged for exclusion due to insufficient specificity: (b) demonstrates low image specificity \enti, (c) illustrates low text specificity \entt, and (a) indicates low specificity in both aspects. Conversely, Figure (d) presents a scenario with high \enti and \entt but low \cossim, highlighting a misalignment between the image and text, evidenced by the absence of an ``elephant print''.}
    \label{fig:overview}
\end{figure}

Recent studies have shown that a machine learning model performance is highly correlated to the training dataset scale and the dataset quality; carefully human-validated high-quality training data leads to a better model performance than the same size of noisy data \cite{should_clip,chinchilla}.
However, human-validated dataset construction is labor-intensive, making its scale-up expensive and impractical.
As an alternative, there have been attempts to scale up noisy data points until reaching the performance garnered from carefully collected high-quality training datasets \cite{openclip,align,basic}.
However, this approach requires more than billion-scale data points that introduces another challenges in computational costs and storage size.
To mitigate the problem, researchers have begun to study inexpensive automatic data filtering approaches to the noisy billion-scale data points.

The large datasets being created today, except private in-house datasets \cite{jft300m,jft3b,ig1b,hqitp}, rely heavily on web-crawled documents by CommonCrawl.
As the scale of images and texts obtained from the web is tremendously large, each dataset employs different heuristics for reducing the size of the dataset.
These heuristics include, for example, whether the text is a title from Wikipedia, whether it is written in English, and whether the image resolution is large enough \cite{clip,metaclip,laion,laion400m,datacomp}.
Another rule-of-thumb is model-based filtering, usually based on the pre-trained CLIP \cite{clip} model, which determines if the given image and text are semantically aligned \cite{laion,laion400m,datacomp}, or if the given image is similar to high-quality images from human-validated datasets, such as ImageNet \cite{datacomp}.

Although CLIP-based filtering helps verify the semantic \emph{alignment} between images and texts, we argue that \emph{alignment} alone is insufficient for high-quality data filtering.
More specifically, we should consider \emph{specificity} of each data point.
Here, we (informally) define \emph{alignment} as whether a given image-text pair is sufficiently similar and \emph{specificity} as whether a given unimodal data point contains sufficient information to be uniquely defined (\ie, specificity measures how each data point has semantically overlapping with other data points). A more formal definition will be described in \cref{sec:econe}.
\cref{fig:overview} illustrates the concept of alignment and specificity. In the figure, the website screenshot and the URI are well-aligned, but the information of the screenshot and the URI are not sufficiently enough to be uniquely defined. Unfortunately, as CLIP-based filtering only considers alignment, it cannot filter out underspecified images and texts.

To consider both \emph{alignment} and \emph{specificity}, we employ the pre-trained CLIP \cite{clip} and its hyperbolic embedding version, MERU \cite{meru}.
By employing both alignment and specificity metrics, our data filtering, named HYPerbolic Entailment filtering (\ours), can successfully handle underspecified samples and misaligned pairs at the same time.
More specifically, we propose a novel specificity measurement based on the property of hyperbolic embeddings, the image specificity \enti and the text specificity \entt.
We employ four metrics for \ours: the cosine similarity (\cossim) between the two CLIP embeddings, the negative Lorentzian distance (\negl) \cite{intro_textbook} between the two MERU embeddings, and the specificity measure of each modality using the entailment cone defined by MERU: \enti (how specific the image is) and \entt (how specific the text is). \ours utilize all four metrics: \enti, \entt, \negl, and \cossim for filtering, making sure that the samples like shown in \Cref{fig:overview} are eliminated, which is not possible by alignment-only filtering.
By considering various aspects of data points rather than only alignment, \ours is ranked in the first place on the Datacomp filtering track \cite{datacomp} for small and medium scales by combining with DFN \cite{dfn}.
Our contribution can be summarized as follows.
\begin{enumerate}
\item We propose \ours, a novel method that enhances the training of CLIP models beyond what is possible with traditional CLIP-based filtering techniques by leveraging uni-modal \emph{specificity} along with cross-modal \emph{alignment}.
\item \ours can be effectively used independently or in conjunction with other filtering methods. When combined, it achieves a new state-of-the-art in the DataComp benchmark, indicating its ability to filter datasets using distinct properties compared to other methods.
\item \enti can be independently applied to induce a dataset for training image-only self-supervised models, showing superior performance compared to alignment-based filtering.
\end{enumerate}

\section{Background}
\label{sec:background}

\subsection{Hyperbolic Embeddings}

Despite the usefulness of Euclidean embeddings, they cannot capture additional instance-wise information, such as specificity. In this paper, we employ hyperbolic embeddings to capture additional information for data filtering.
A hyperbolic space maps data that needs to be close to many positives at the same time (\ie, more generic data) into closer to the origin, while maps data with fewer positive pairs (\ie, more specific data) into farther away from the origin \cite{intro_textbook, poincare}. Conceptually, the distance from the origin corresponds to the uncertainty represented by Euclidean Gaussian embeddings \cite{tifrea2018poincareglove}.
Thus, hyperbolic embeddings can naturally capture how the uncertainty of inputs caused by inherent noisy image-text pairs \cite{pcmepp}.
Practical implementations of $\mathbb{R}^n$ hyperbolic spaces include the Poincaré ball model \cite{poincare,hetero_poincare,econe,vecone,hie,his,hypervit,hypercl}, which distorts the distances in $\mathbb{R}^n$, and the hyperboloid model (Lorentz model), which is defined as a sub-manifold of $\mathbb{R}^{n+1}$ \cite{lorentzcone,meru}.
A recent study, MERU \cite{meru}, has successfully extended this concept to image-text contrastive models, showing better performance than CLIP in cross-modal retrieval and illustrating interesting applications of image traversal.
In this paper, we focus on noisy pair filtering leveraging the \emph{specificity} we can gather from the hyperbolic model, which was not addressed in MERU, and show the advantages of using hyperbolic CLIP as a filtering network.
To be self-contained, we will describe the details of hyperbolic embeddings and how specificity can be measured by hyperbolic embeddings in \Cref{sec:hyperbolic_clip}.

\subsection{DataComp Benchmark}
\label{sec:datacomp}

For recent years, several evaluation benchmark suites have been proposed to evaluate \textit{models} on various modalities, including text \cite{glue,superglue}, images \cite{vtab}, video \cite{value}, and multimodal models \cite{vlue,slue}.
These model-driven benchmarks include evaluation datasets and tasks, but they do not limit models and training datasets.
Namely, the three factors of the scaling law \cite{scaling_law,beyond_scaling_law,chinchilla,transfer_scaling_law} --the size of the model, the amount of data, and the number of training steps-- cannot be controlled through the benchmarks.
It makes fair quantitative comparisons between different algorithms or methods difficult beyond the effect of scale.
To address this, DataComp \cite{datacomp} has been proposed as a data-driven, rather than model-driven, benchmark where the size of the model and the number of training steps (the number of samples seen) are controlled and fixed.
The Datacomp evaluation consists of 38 tasks, mainly grouped into four task groups: ImageNet, 6 ImageNet distribution shifts \cite{imagenet_a,imagenet_r,imagenet_v2,objectnet}, 13 VTAB \cite{vtab}, and 3 retrievals \cite{coco,flickr30k,winogavil}.
The main evaluation metric of DataComp is computed by the average score of these tasks, and additional benchmarks from CLIP \cite{clip} and WILDS \cite{wilds}.
In this paper, we consider the \textbf{DataComp filtering track}, a benchmark for evaluating the effectiveness of filtering methods.
There are four different scales of datasets in terms of fixed model size, training budget, and the number of seen samples  (\texttt{small}, \texttt{medium}, \texttt{large}, and \texttt{xlarge}). For example, the number of seen samples of \texttt{small} is 12.8M, growing 10 times for each scale (\eg, \texttt{large} has 1.28B seen samples).
Therefore, for each filtering track, the training method, budget, and the number of seen samples are fixed, but only the seen samples are changed.
We note that the training method is fixed as CLIP training and the evaluation protocol is fixed as the average zero-shot score on the 38 tasks of Datacomp evaluation suite.
It is because CLIP demonstrates a better scaling trade-off than other methods \cite{should_clip,metaclip}, and there exist well-founded open software \cite{openclip,openclip_software} and open datasets \cite{nguyen2022quality,laion400m,coyo,laion} for the training.
\section{Method}
This section outlines the overview of HYPE filtering, the theoretical background, and the practical implementation of hyperbolic embeddings, presenting \ours as an effective method for dataset filtering in image-text contrastive learning.

\subsection{Overview of HYPE}

\begin{figure}[t]
    \centering
    \includegraphics[width=\linewidth]{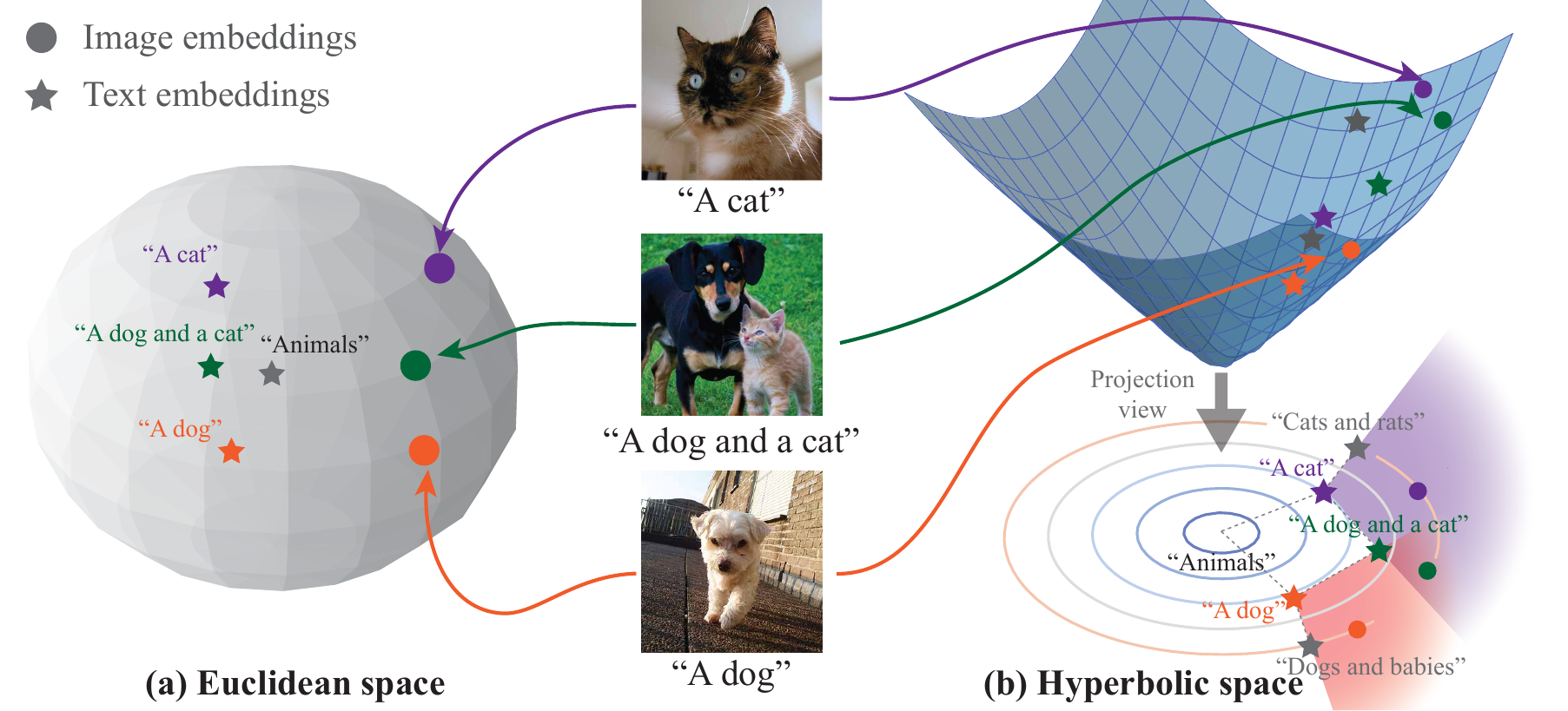}
    \caption{Conceptual comparisons of Euclidean embeddings and hyperbolic embeddings.}
    \label{fig:euclidean_vs_hyperbolic}
\vspace{-1em}
\end{figure}

While CLIP-based filtering captures \emph{alginment} well, it cannot effectively measure the \emph{specificity} of each data point.
More specifically, as CLIP is trained with noisy-contrastive estimation \cite{nce,cpc} using random samples as negative pairs, CLIP enforces to make each embedding located to a more distinct subspace rather than having semantic overlaps between each other.
For example, consider a photo of a dog and a cat together and captions \emph{``A dog''}, \emph{``A cat''}, and \emph{``A dog and a cat''} in \Cref{fig:euclidean_vs_hyperbolic}.
In this case, as shown in \Cref{fig:euclidean_vs_hyperbolic} (a), the best Euclidean space mapping will map the dog and cat photo to the midpoint between the embedding of \emph{`A dog'} and \emph{`A cat'}, because the dog and cat photo should be matched with both dog and cat embeddings.
However, the actual semantic meaning is more complex than the average of the two embeddings.
As pointed out by \citet{meru}, it is because CLIP uses the same distance metric at every point.

Hyperbolic embedding, on the other hand, can capture more complex semantics by letting each point have different distance metrics.
As shown in \Cref{fig:euclidean_vs_hyperbolic} (b), hyperbolic embedding space can represent more complex information than Euclidean embedding space.
Conceptually, a more generic data point (\ie, potentially matched with more samples) will be mapped into a point close to the center point in hyperbolic space.
For example, the textual embeddings of \textit{``A cat''} and \textit{``A dog''} are closer to the center (\textit{``Animals''}) than that of \textit{``A dog and a cat''} and \textit{``Cats and rats''}.
Moreover, using the property of hyperbolic embedding space, we can define an ``entailment'' of each modality, \ie, whether the given data sample can be matched with the other data samples.
For example, \Cref{fig:euclidean_vs_hyperbolic} (b) also illustrates the projected view of the hyperbolic space.
In the projected view, we can observe that the ``A dong and a cat'' caption embedding is placed where the ``cones'' of caption embeddings ``A cat'' and ``A dog'' (shown in purple and red areas, respectively) are overlapped.
In other words, by using the concept of the ``entailment cone'', we can define the entailment of the given input.

\begin{figure}[t]
\centering
\includegraphics[width=\linewidth]{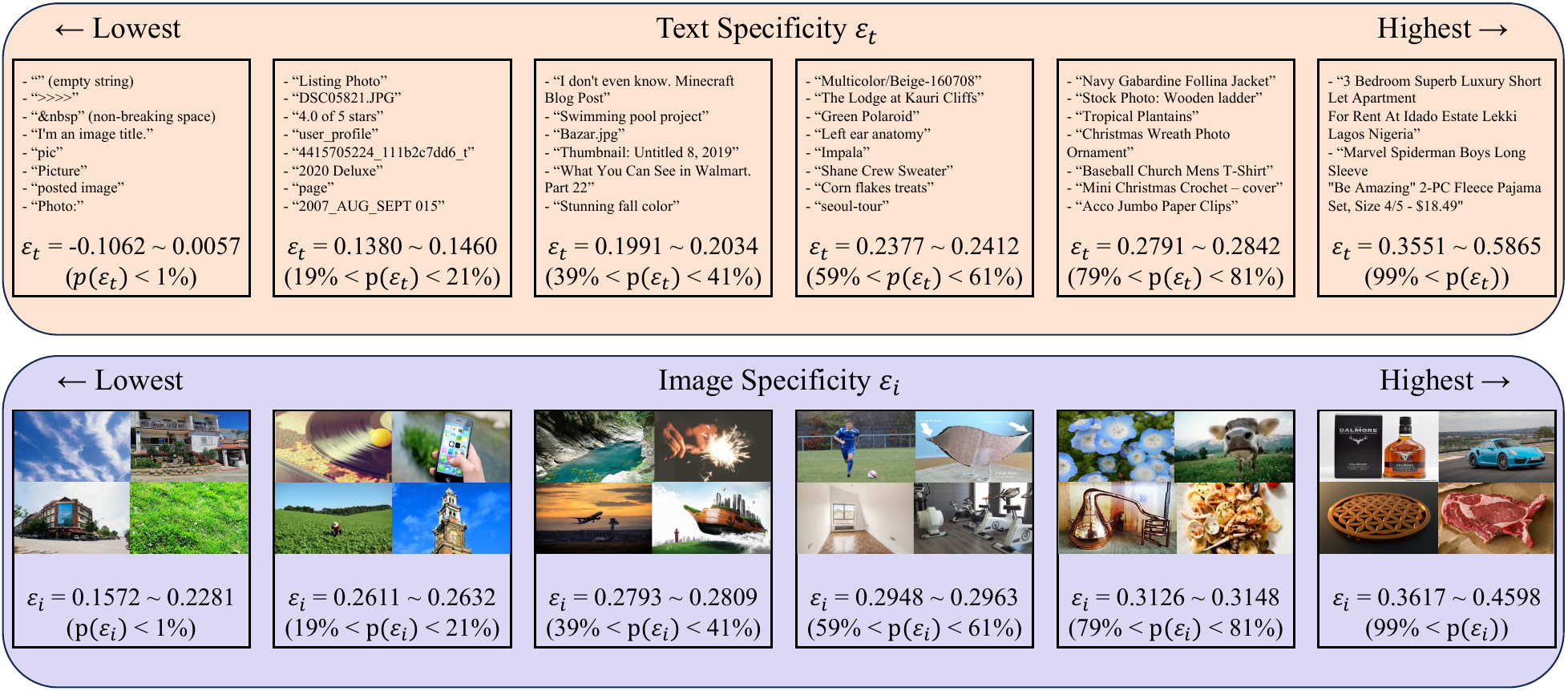}
\caption{
We show examples of low and high \enti and \entt from the 12.8M Datacomp small pool, where each percentile group spanned with 20\% intervals.
Here, a higher value denotes that the instance is more specific (see \Cref{sec:econe} for details of \enti and \entt).
The range absolute value and their percentile $p(\cdot)$ of \enti and \entt are also shown.
For texts, the lowest \entt texts are empty sentences or the least specific texts that could fit any image, such as \emph{``Picture''}, while the higher \entt texts are generally longer sentences that describe some object in detail.
For images, images with low \enti are either background images with no objects or with too many objects, while images with higher \enti are so-called \emph{iconic} images, which contain a single object that can be described with precision.
}
\label{fig:entailment_and_specificity}
\vspace{-1em}
\end{figure}

Using the entailment cones, we define the ``entailment loss'' $\mathcal{L}_{e}(\vx, \vy)$ for the given image-text pair that measures whether the image $\vy$ is correctly placed on the entailment cone of the corresponding text $\vx$.
Then, we define the ``specificity'' of each input by computing the average entailment loss on the dataset.
The image specificity \enti is defined as the average entailment loss, \ie, $\sum_\vx \frac{\mathcal{L}_{e}(\vx, \vy)}{M}$, and the text specificity \entt is defined similarly, $\sum_\vy \frac{\mathcal{L}_{e}(\vx, \vy)}{M}$.
\enti and \entt measure whether the learned hyperbolic embedding space describes the given input well.
We will provide a more rigorous mathematical definition in the latter section.
\Cref{fig:entailment_and_specificity} shows examples of images and texts with low and high specificity values (\ie, \enti and \entt, respectively).
As shown in the figure, samples with smaller specificities are more generic and underspecified.
For example, the low \enti values of mobile phone or tower images denote their abundant potential relative captions in the dataset.
Conversely, Dalmore whisky in the ``Highest'' category highlights the scarcity of descriptive texts without directly mentioning ``Dalmore'', underscoring the metric's effectiveness in distinguishing specificity.
Similarly, the captions \textit{``pic''} and \textit{``Picture''} have low \entt values as they are vague to describe a specific image.

In this paper, we propose to use not only CLIP alignment score, \cossim, but the specificity scores \enti and \entt.
Also, as the CLIP embedding space is not sufficient to represent complex image-text representations (as shown in \Cref{fig:euclidean_vs_hyperbolic}), we use the alignment score measured by our hyperbolic CLIP, \negl.
Finally, following the baseline DataComp filtering, we additionally employ the ImageNet clustering filter $c_{\text{IN}}$, which denotes whether the given image belongs to ImageNet classes.
Our \ours score is defined as follows:
\begin{equation}
\label{eq:hype}
\text{HYPE}_{\text{score}} = \epsilon_i + \epsilon_t - d_\mathcal{L} + cos(\theta) + c_{\text{IN}}
\end{equation}

In the following subsections, we will provide the details of the hyperbolic CLIP \cite{meru} and more formal theoretical explanations of the meaning of \enti and \entt.

\subsection{Hyperbolic CLIP}
\label{sec:hyperbolic_clip}

In this subsection, we provide a brief introduction to hyperbolic embeddings and its multimodal version, MERU \cite{meru}.
Hyperbolic embeddings have been actively studied on diverse modalities, such as images \cite{hcl} or text \cite{valentino2023multirelational}.
Recently, \citet{meru} applied hyperbolic embeddings to image-text joint embedding space based on CLIP, named MERU.
MERU is based on the Lorentz model, which uses the upper half of a two-sheeted hyperboloid in the $n+1$-dimensional Euclidean space $\bbR^{n+1}$ to represent the $n$-dimensional hyperbolic space $\mathcal{L}^n$.
The $\vx \in \bbR^{n+1} = [\vx_{space}, x_{time}]$ in this space consists of two components \cite{minkowski}:
One is $\vx_{space} \in \bbR^n$ in the $n$-dimensional \emph{space} dimension and the other is $x_{time} \in \bbR$ in the one-dimensional \emph{time} axis.
This hyperboloid is symmetric with respect to the time axis and has a \emph{Lorentzian inner product} $\langle \vx , \vy \rangle_\mathcal{L} = \langle \vx_{space},\vy_{space} \rangle - x_{time} \; y_{time}$, which is different from the Euclidean inner product.
From this inner product, the \emph{Lorentzian norm} is $\lVert \vx \rVert _\mathcal{L} = \sqrt{\lvert \langle \vx,\vx \rangle_\mathcal{L} \rvert}$ is derived.
Since the Lorentz model is defined to have a constant curvature of $-c$ at all points: $\mathcal{L}^n = \{\vx \in \bbR^{n+1} : \langle \vx,\vx \rangle_\mathcal{L} = \sfrac{-1}{c} \; , \; c > 0\}$, we can derive $x_{time}$ from $\vx_{space}$:
\begin{equation}
x_{time} = \sqrt{\sfrac{1}{c} + \lVert \vx_{space} \rVert^2}
\end{equation}

MERU is built upon the Lorentz model and the CLIP architecture.
MERU does not $L^2$ normalize $\vv_{enc} \in \bbR^n$, the embedding that passed the last linear projection in CLIP.
Instead, MERU uses $\vv_{space}=\vv_{enc}$ to define $\vv = [\vv_{enc}, 0] \in \bbR^{n+1}$ and uses it as a point in the tangent space on the hyperboloid origin $\vO = [\mathbf{0}, \sqrt{\sfrac{1}{c}}]$ (this is because $\langle \vO, \vv \rangle _\mathcal{L} = 0$ holds).
MERU multiplies $\vv$ by a learnable scalar $\alpha$ initialized as $\sqrt{\sfrac{1}{n}}$.
The \emph{negative Lorentzian distance}, which we will use as a similarity for the contrastive learning is defined as:
\begin{equation}
-d_\mathcal{L} (\vx, \vy) = -\sqrt{\sfrac{1}{c}} \cdot \cosh^{-1} ( -c \; \langle \vx,\vy \rangle_\mathcal{L} )
\end{equation}
As $-d_\mathcal{L}$ can only be calculated on a manifold, not the tangent space, we need to map $\vv$ in the tangent space to the manifold.
Luckily, as MERU only deals with the tangent space of the origin $\vO$, this \emph{exponential map} can be simplified into:
\begin{equation}
\vx_{space} = \frac{\sinh(\sqrt{c} \; \lVert \vv_{space} \rVert)}{\sqrt{c} \; \lVert \vv_{space} \rVert} \vv_{space}
\end{equation}

By applying the exponential map to text and image embeddings, MERU can find the \negl between positive and negative pairs in a batch, which can be simply used instead of the cosine similarity of CLIP's InfoNCE loss to train the model.
MERU simplifies the exponential map by using the tangent space of the origin, thus minimizing potential numerical instability in the model's computation.

\subsection{Entailment Cone and Specificity}
\label{sec:econe}

Now, we describe how we can measure specificity using hyperbolic embeddings.
Note that \negl also can perform as a filtering metric as a better alignment measure rather than the vanilla CLIP distance \cossim.
However, \negl can only measure \textit{alignment} between images and texts but cannot measure how each image or text is \textit{specific}.
This paper proposes a new instance-wise filtering metric named \textit{specificity} based on the concept of \emph{entailment}.
The concept of \emph{entailment} has its roots in logic and linguistics, long before its incorporation into machine learning \cite{logic1,logic2}.
In logic, entailment is a fundamental relationship where the truth of one statement guarantees the truth of another.
In natural language processing (NLP), a number of tasks have been created to verify that the language model can properly capture this entailment relationship (\ie, semantic containment and exclusion): RTE \cite{rte1,rte2,rte3,rte4}, MNLI \cite{mnli}, WNLI \cite{wnli}, etc., and these tasks form a significant part of the GLUE benchmark \cite{glue,superglue}.
Beyond NLP, tasks have also been created in the vision-and-language domain, such as SNLI-VE \cite{snlive1,snlive2}, to evaluate cross-modal entailment relationships between images and text.
The concept of an \emph{entailment cone} emerges when we consider how entailment relationships can be represented in a vector space.
The idea is that for a given concept represented by a vector, there exists a \emph{cone} in the vector space within which all vectors that are semantically entailed by the original term fall.

\begin{wrapfigure}{r}{0.25\linewidth}
\vspace{-2em}
\includegraphics[width=\linewidth]{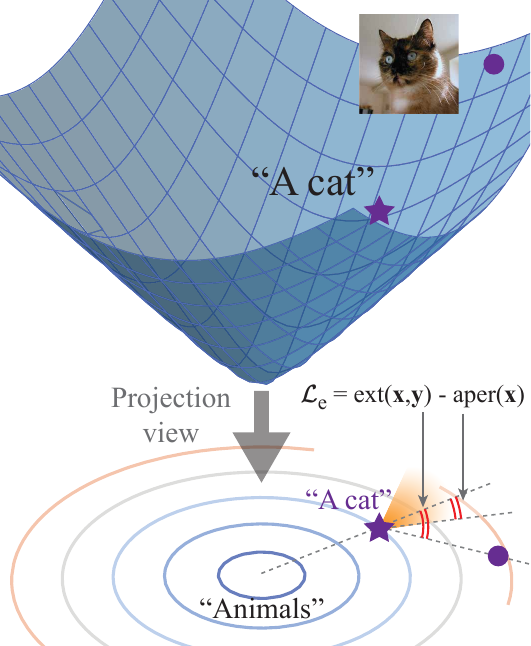}
\caption{Visual example of aper \cref{eq:aper}, ext \cref{eq:ext} and entailment loss \cref{eq:le}.}
\label{fig:loss_vis}
\vspace{-1em}
\end{wrapfigure}
While the implementation of entailment cones in the vision-and-language context can be done through order embedding \cite{order_embedding}, \citet{meru} borrows the concepts of \citet{econe} and \citet{lorentzcone} to train MERU using entailment loss, which is involved in the training of the model.
In the hyperboloid space drawn by MERU, an entailment cone is defined as a half-aperture with $K=0.1$:
\begin{equation}
\label{eq:aper}
\text{aper}(\vx) = \sin^{-1} \left( \frac{2K}{\sqrt{c} \; \lVert \vx_{space} \rVert} \right)
\end{equation}
\citet{meru} empirically demonstrated that \emph{text always entails an image}.
This concept can be taken for granted because text, with its symbolic representation, is always less specific than an image with pixel-level specificity.
Thus, entailment loss makes the model learn such that the image embedding of a positive image-text pair falls within the cone of its paired text (See \Cref{fig:euclidean_vs_hyperbolic} (b) as an example).
The acute angle that the image embedding $\vy$ makes with the text embedding $\vx$ can be found following hyperbolic trigonometry:
\begin{equation}
\label{eq:ext}
\text{ext}(\vx, \vy) = \cos^{-1} \left( \frac{y_{time} + x_{time} \; {c \; \langle \vx,\vy \rangle _\mathcal{L}}}{\lVert \vx_{space} \rVert \sqrt{\left( {c \; \langle \vx,\vy \rangle _\mathcal{L}} \right)^2 - 1}} \right)
\end{equation}
Entailment loss is then determined by the difference between this deviation and the size of the cone:
\begin{equation}
\label{eq:le}
\mathcal{L}_{e}(\vx, \vy) = \max(0, \; \text{ext}(\vx, \vy) - \text{aper}(\vx))
\end{equation}

The visual explanation of \Cref{eq:aper}, \ref{eq:ext} and \ref{eq:le} is illustrated in \Cref{fig:loss_vis}.
The $\mathcal{L}_{e}$ alone still requires image-text pairs.
To use this value independently to measure uni-modal specificity, we first sorted all samples from the DataComp medium in descending order of CLIP similarity, and then selected the top $N$ samples.
We then measured the $\mathcal{L}_{e}$ for each image and text MERU embedding in the DataComp medium against the MERU embeddings of the opposite modality in the $N$ samples and averaged these values.
We used the $M$ images and $M$ texts with the highest average $\mathcal{L}_{e}$ as our reference set: $\mathcal{S}_i$ and $\mathcal{S}_t$, respectively.
We set $N$ and $M$ to 20,000 as the value of \ent converged when calculated over 3,000 samples.
The relatively low variance of metrics shown in \Cref{tab:metric_stat} shows that the specificity values remain consistent across different reference sets, suggesting that it is invariant to the choice of dataset and not subject to bias.
Now, given any image, we can calculate its $\mathcal{L}_{e}$ with the $M$ text reference set, and we define this value as image specificity \enti.
Similarly, we can calculate the $\mathcal{L}_{e}$ value for text, and define this value as text specificity \entt:
\begin{equation}
\epsilon_{t}(\vx) = \sum_{\vy \in \mathcal{S}_i} \sfrac{\mathcal{L}_{e}(\vx, \vy)}{M}\; \text{and}\; \epsilon_{i}(\vy) = \sum_{\vx \in \mathcal{S}_t} \sfrac{\mathcal{L}_{e}(\vx, \vy)}{M}
\end{equation}

\begin{table}[t]
\setlength{\tabcolsep}{3pt}
\centering
\caption{\textbf{DataComp statistics.} We have fewer samples than the original release of DataComp small (12.8M) and medium (128M) due to inaccessible URLs.
We confirmed that the overall metric statistics of the samples remain largely unchanged for both scales.
Hence, we expect that using these metrics as filtering will achieve almost similar results even when the scale goes beyond DataComp medium.
Also, \entt is significantly lower than \enti, namely, \emph{text always entails an image} empirically.
}
\vspace{-1em}
\begin{adjustbox}{width=\textwidth,center}
\begin{tabular}{@{}llccccc@{}}
\toprule
Dataset & Size & \entt & \enti & \negl & \cossim & $c_{\text{IN}}$ \\ \midrule
DataComp Small & 11.6M & $0.211 \pm 0.082$ & $0.289 \pm 0.030$ & $-0.726 \pm 0.053$ & $0.208 \pm 0.064$ & $6.110 \pm 4.875$ \\
DataComp Medium & 115.6M & $0.210 \pm 0.082$ & $0.289 \pm 0.030$ & $-0.726 \pm 0.053$ & $0.208 \pm 0.064$ & $5.957 \pm 4.908$ \\ \bottomrule
\end{tabular}
\end{adjustbox}
\label{tab:metric_stat}
\vspace{-1em}
\end{table}

\subsection{Hyperbolic Entailment Filtering (\ours)}
\label{sec:epsilon}
Here, we describe the details of \ours.
We first train a MERU model with ViT-B/16 and ViT-L/14 backbones on CC3M \cite{cc3m} and CC12M \cite{cc12m} in addition to RedCaps \cite{redcaps}.
Both models were trained on 8 V100s with a batch size of 2048.
The models were optimized using AdamW \cite{adamw}, with a weight decay of 0.2, $(\beta_1, \beta_2)=(0.9, 0.98)$, and a learning rate of $5 \times 10^{-4}$.
After a warm-up of 4,000 steps, ViT-B/16 was trained for 62,500 steps and ViT-L/14 for 125,000 steps using a cosine decay learn rate schedule.
Our implementation is based on the OpenCLIP codebase \cite{openclip_software}.
Training of ViT-B/16 and ViT-L/14 MERU models took approximately 10 hours and 61 hours, respectively.

\begin{table}[t]
\setlength{\tabcolsep}{10pt}
\centering
\caption{ImageNet-1k \cite{imagenet} zero-shot classification accuracy (IN1K) and MS-COCO \cite{coco} text-to-image (T2I) and image-to-text (I2T) retrieval recalls on Karpathy test split \cite{coco_karpathy} and mAP on ECCV Caption \cite{eccv_caption} performances of CLIP and MERU models. Note that the results reported in \citet{meru} used COCO 2017 validation split instead of Karpathy test split. The results marked with an asterisk ($\ast$) are the official checkpoints from \cite{meru}, and the unmarked ones are the ones we reproduced. The best scores are in \textbf{bold} and the second best scores are in \underline{underlined}.}
\vspace{-1em}
\begin{adjustbox}{width=\textwidth,center}
\begin{tabular}{@{}llccccccccccc@{}}
\toprule
\multirow{2}{*}{Model} & \multirow{2}{*}{Method} & \multirow{2}{*}{\shortstack{Dataset\\Size}} & \multirow{2}{*}{\shortstack{\# Samples\\Seen}} & \multirow{2}{*}{IN1K} & \multicolumn{4}{c}{COCO T2I} & \multicolumn{4}{c}{COCO I2T} \\
 &  &  & & & R1 & R5 & R10 & mAP & R1 & R5 & R10 & mAP \\ \midrule
B/16 & CLIP $\ast$ & 12M & 245M & 37.9 & 15.4 & 34.3 & 44.4 & 18.5 & 21.2 & 43.4 & 54.1 & 10.3 \\
 & MERU $\ast$ & 12M  & 245M & 37.5 & 15.1 & 33.8 & 44.8 & 18.6 & 21.2 & 43.0 & 53.9 & 10.0 \\
 & MERU & 27M & 128M & \underline{42.3}   & \underline{24.6} & \underline{49.0}  & \underline{60.8} & \underline{28.8} & \underline{37.9} & \underline{63.4}  & \underline{75.0} & \underline{18.3} \\
L/16  & CLIP $\ast$ & 12M  & 245M & 38.4 & 14.2 & 32.1  & 42.6 & 17.6 & 21.2 & 41.9  & 52.2 & 9.8 \\
 & MERU $\ast$ & 12M  & 245M & 38.8 & 14.7 & 33.2  & 43.4 & 18.5 & 21.2 & 42.1  & 52.7 & 10.2 \\
L/14 & MERU & 12M  & 128M & 38.2 & 13.6 & 31.2 & 41.0 & 17.6 & 21.2 & 44.2 & 54.6 & 10.3 \\
  & MERU & 27M & 256M &  \textbf{50.2} & \textbf{30.2} & \textbf{55.4}  & \textbf{66.9} & \textbf{32.8} & \textbf{43.3} & \textbf{69.5}  & \textbf{79.7} & \textbf{21.0} \\ \bottomrule
\end{tabular}
\end{adjustbox}
\label{tab:meru_repro}
\end{table}

Note that the original MERU by \citet{meru} was trained solely on the RedCaps \cite{redcaps} dataset.
We added more clean data points to allow better filtering capability, as the findings of DFN \cite{dfn} and our discussion in \cref{sec:datacomp_baselines}.
We also note that the original MERU uses ViT-B/16 and ViT-L/16 backbones with their textual encoder having a hidden size of 512.
Since DataComp \cite{datacomp} uses ViT-B/16 and ViT-L/14 for its baseline CLIP filtering method, we retrained MERU on ViT-B/16, which has a 512 textual encoder hidden size, and ViT-L/14, which has a 768 textual encoder hidden size, with the expanded dataset.
The results of MERU re-training are shown in \Cref{tab:meru_repro}.
Surprisingly, even when all the training hyperparameters, including the batch size, were the same as in the original MERU, and the training was done with fewer steps (ViT-B/16), the zero-shot performance of the reproduced MERU model was significantly better than that of the original MERU.
All results in this paper are based on the hyperbolic embeddings obtained by our reproduced ViT-L/14 MERU.

We extract \enti, \entt, and \negl for every sample in the target image-text dataset using our MERU model.
For each sample, we also compute and store the ImageNet clustering-based image filter used by DataComp and the CLIP score \cossim of the ViT-L/14 CLIP.
The clustering-based filter $c_{\text{IN}}$ is quantified as a value of 10 if included and 0 if not, enabling preferential use.
\Cref{tab:metric_stat} summarizes the statistics for the datasets tested in this paper.
The \ourss is obtained by linearly combining all the metrics with equal weight as defined in \Cref{eq:hype}.

Note that the metrics used for \ours have the same computation complexity as the CLIP distance.
On the other hand, a number of the existing filtering methods need more complex computations, such as the OCR engine (T-MARS \cite{tmars}) and additional clustering operations (CIDS \cite{devil}).
Also, we argue that our method is data-efficient compared to the previous model-driven filtering methods (our method: 27M, CLIP: 400M, DFN \cite{dfn}: 2B)
Our method is simple yet archives the first place in small and medium DataComp leaderboards.

\subsection{Ablation study}
\begin{figure}[t]
    \centering
    \includegraphics[width=\linewidth]{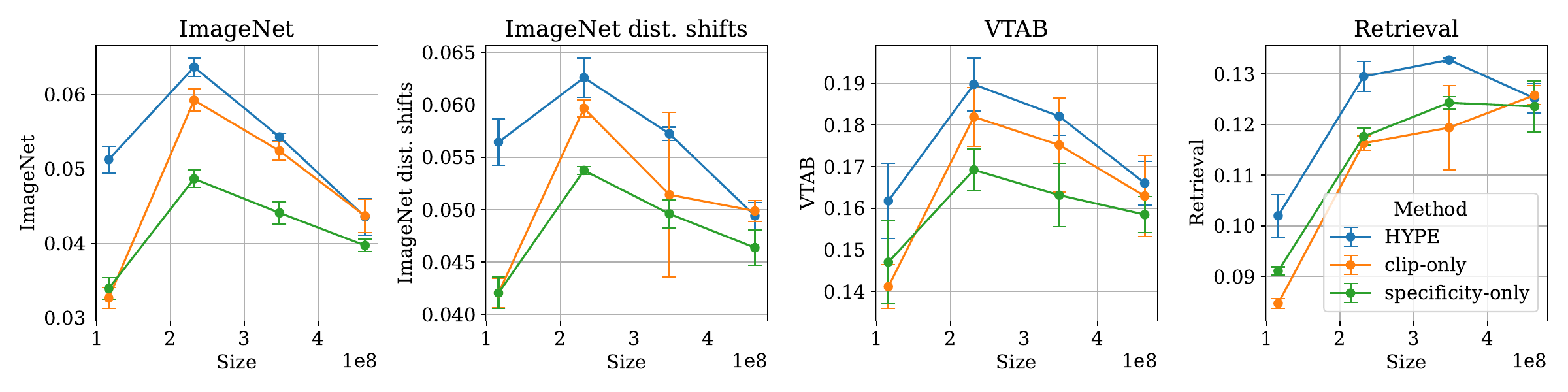}
    \vspace{-2em}
    \caption{\textbf{Comparisons with baseline filtering methods and \ours.} We show the subsampled Datacomp training set from 10\% to 40\% and evaluate them across four Datacomp benchmark task groups. Each model was trained four times with varied seeds.
    10\% and 30\% results are the same as \Cref{tab:datacomp}.}
    \label{fig:metric_composition}
\end{figure}
In this subsection, we provide ablation studies of \ours design choices.
First, we show that using our metric outperforms solely using CLIP similarity or solely using specificity in \Cref{fig:metric_composition}.
Across sample sizes from 10\% to 40\% and across four Datacomp benchmark task groups, \ours consistently outperformed each metric alone.
Note that the gaps can be small in 40\% samples because they share more samples, thus less filtering effect.
In 10\% or 20\%, where filtering works more sensitively, \ours always outperforms the baseline methods with large gaps.
This demonstrates that, as suggested in \Cref{fig:overview}, each metric, when used in isolation, is limited in its ability to filter out data that adversely affects image-text contrastive learning.

\begin{wraptable}{r}{.5\linewidth}
\vspace{-4em}
\centering
\caption{Ablation study}
\label{tab:abl}
\begin{adjustbox}{width=\linewidth,center}
\begin{tabular}{@{}lcccc@{}}
\toprule
Method & IN & VTAB & Ret & Avg  \\ \midrule
\rowcolor{Gray}
\ours & \textbf{0.338} & 0.357 & \textbf{0.286} & 0.343 \\
\ours $-$ $c_{\text{IN}}$ & 0.322 & \textbf{0.369} & 0.273 & \textbf{0.349} \\
\ours $-$ $c_{\text{IN}}$ $-$ \cossim & 0.320 & 0.358 & 0.278 & 0.345 \\
\cossim only \cite{datacomp} & 0.260 & 0.326 & 0.235 & 0.322 \\
\cossim $+$ $c_{\text{IN}}$ \cite{datacomp} & 0.297 & 0.346 & 0.231 & 0.328 \\
\bottomrule
\end{tabular}
\end{adjustbox}
\vspace{-2em}
\end{wraptable}
We also examined the effect of each component of \ours in \Cref{tab:abl}.
Our findings confirm that while $c_{\text{IN}}$ enhances IN zero-shot, omitting $c_{\text{IN}}$ yielded superior average performance (1st vs. 2nd rows). 
The results of removing \cossim (3rd row) are inspiring: our model is trained with 1/15 data points and 1/5 seen training samples than OpenAI CLIP but performs better than the CLIP baseline (4th row).
We also found that 
there is no single weight combination for \Cref{eq:hype} that performs best for all datasets.
In this paper, we set all weights as 1 (\ie, 1st row), considering the importance of the ImageNet benchmark and the relatively low importance of small datasets, such as SVHN in the VTAB benchmark.
\section{Experiments}

In this section, we will show and discuss the results of using \ours for the image-text contrastive learning benchmark DataComp's small and medium, and \enti for image-only contrastive learning by itself.
Before that, we will discuss the methods we used as a baseline for filtering in image-text contrastive learning.

\subsection{Comparison Methods}
\label{sec:datacomp_baselines}

In this paper, ``filtering'' refers to the process of excluding samples from the training data, while ``sampling'' refers to how often each sample is used for training.
Here, we introduce the major baselines of the DataComp filtering benchmark.
The most simple baseline filters the dataset by language (\eg, leaving only English text), text length (\eg, more than two words or five characters), and image size (\eg, aspect ratio of 3 or less and shorter axis more than 200 pixels). 
There are two methods that empirically perform well.
One is image-based clustering, which groups the CLIP embeddings 100K centroids and then filters the samples in centroids based on whether one of the images in ImageNet is closest to the centroid of the cluster to which each sample belongs.
The other is CLIP score filtering we explained before.
Recently, three notable approaches have been proposed for Datacomp medium scale: DFN \cite{dfn}, CIDS \cite{devil}, and T-MARS \cite{tmars}.

\noindent\textbf{Data Filtering Networks (DFN)} \cite{dfn} is a model-centric approach without multi-step filtering; they directly train a network determines whether filtering out the given data.
The authors showed that CLIP cosine similarity-based DFN performs the best among the other possible variants, such as, autoencoder \cite{m3ae}.
The DFN paper also observes that training DFN with high-quality training samples (\ie, a proprietary dataset, such as HQITP-357M \cite{hqitp,dfn}) is crucial for better filtering, compared to low-quality and large-scale training samples.
Note that the best-performing DFN is trained on HQITP-357M, whose scale is already beyond the DataComp medium of 128M, making it very resource-intensive.

\noindent\textbf{Cluster-Importance-based Data Selection (CIDS)} \cite{devil} uses the 38 Datacomp evaluation datasets to filter out samples dissimilar to the target evaluation datasets and then duplicates the samples with similar distributions for more extensive training sampling.
While this method does not require a significant amount of additional computing resources, it has a notable drawback: the model needs to know on which dataset the CLIP will be evaluated before filtering.

\noindent\textbf{Text-Masking and Re-Scoring (T-MARS)} \cite{tmars} reveals that many samples in noisy datasets, like DataComp's dataset pool, are simple OCR samples (image-text pairs that simply transcribe the text in the image).
This helps CLIP focus on learning visual semantics by retaining only those images in the data pool that still have high CLIP scores after masking the text in the images.
However, removing all OCR-like samples would harm the performance of tasks like MNIST \cite{mnist}, SVHN \cite{svhn}, and RenderedSST2 \cite{clip} in DataComp's evaluation dataset; therefore, they still require CLIP to read the text in the images.

\subsection{Image-Text Contrastive Pre-training}
\begin{table}[t]
\setlength{\tabcolsep}{3pt}
\centering
\caption{We have compared \ours with concurrent works challenging the Datacomp benchmark. Methods with an asterisk ($\ast$) are our reproductions given their sample IDs for a fair comparison, as we were able to download fewer samples than the original models.
\ours on the Datacomp small scale reports values from the average of four models trained with different seeds.
The uniform column stands for whether or not each method uses the given sample IDs with equal probability during training.
The best scores are in \textbf{bold}, and the second best scores are in \underline{underlined}.}
\vspace{-1em}
\begin{adjustbox}{width=\textwidth,center}
\begin{tabular}{@{}lcccccccc@{}}
\toprule
Method & Datacomp Scale & Sample Size & Uniform & ImageNet & ImageNet Dist. Shift & VTAB & Retrieval & Average \\ \midrule
CLIP L/14 30\% \cite{datacomp} & Small & 3.8M & Yes & 0.051 & 0.055 & 0.190 & 0.119 & 0.173 \\
WS \cite{ws_small} & Small & 4.1M & Yes & 0.056 & 0.061 & 0.196 & 0.132 & 0.180 \\
\midrule
\ours 10\% & Small & 1.2M & Yes & 0.051 & 0.056 & 0.162 & 0.102 & 0.150 \\
\ours 20\% & Small & 2.3M & Yes & 0.064 & 0.063 & 0.190 & 0.130 & 0.176 \\
\ours 30\% & Small & 3.5M & Yes & 0.054 & 0.057 & 0.182 & 0.133 & 0.170 \\
\midrule
CLIP L/14 30\% \cite{datacomp} & Medium & 38.0M & Yes & 0.273 & 0.230 & 0.338 & 0.251 & 0.328 \\
WS \cite{ws_small} & Medium & 24.8M & Yes & 0.305 & 0.253 & 0.363 & 0.270 & 0.342 \\ %
T-MARS \cite{tmars} & Medium & 23.0M & No & 0.338 & 0.274 & 0.371 & 0.231 & 0.357 \\
CIDS \cite{devil} $\ast$ & Medium & 21.3M & No & 0.326 & 0.262 & 0.372 & 0.258 & 0.365 \\
DFN \cite{dfn} $\ast$ & Medium & 17.1M & Yes & \underline{0.376} & \underline{0.300} & 0.384 & 0.284 & 0.372 \\
\midrule
\ours 10\% & Medium & 11.6M & Yes & 0.327 & 0.257 & 0.365 & 0.246 & 0.340 \\
\ours 20\% & Medium & 23.1M & Yes & 0.338 & 0.269 & 0.357 & \underline{0.286} & 0.343 \\
\ours 30\% & Medium & 34.7M & Yes & 0.300 & 0.243 & 0.337 & 0.276 & 0.332 \\
\ours 10\% + CIDS \cite{devil} $\ast$ & Medium & 18.9M & No & 0.346 & 0.276 & \underline{0.390} & 0.264 & \underline{0.373} \\
\ours 10\% + DFN \cite{dfn} $\ast$ & Medium & 21.5M & No & \textbf{0.382} & \textbf{0.303} & \textbf{0.393} & \textbf{0.306} & \textbf{0.379} \\
\bottomrule
\end{tabular}
\end{adjustbox}
\label{tab:datacomp}
\end{table}
\Cref{tab:datacomp} includes the DataComp filtering track results of the main competitors (\ie, DFN \cite{dfn}, CIDS \cite{devil} and T-MARS \cite{tmars}) and the ensemble filtering with weak supervision \cite{ws_small}.
As mentioned in \Cref{tab:metric_stat}, we only use the subset of the official DataComp due to the dissipated URLs (about 10\% samples were lost).
For a fair comparison, we obtained the sample IDs used by the two best-performing methods on DataComp medium: CIDS \cite{devil} and DFN \cite{dfn}, and reproduced the model with only those belonging to our pool -- denoted with asterisk ($\ast$).

In the table, we observe that \ours performs extremely well in retrieval scores, \eg, \ours 20\% Medium shows 0.286 retrieval, which outperforms all the baselines.
We believe that it is because hyperbolic embeddings significantly improve retrieval performances compared to the CLIP embedding (as observed in \Cref{tab:meru_repro}), making the filtered data samples by \ours more suitable for retrieval tasks.
This is especially noteworthy given that DFN used 357M high-quality proprietary image-text pairs while \ours is achievable with a much smaller 27M publicly accessible dataset.
Note that DataComp only contains 3 retrieval task groups out of 38 tasks; therefore, if we add more retrieval tasks for the evaluation benchmark, \ours will achieve a higher average score than others.

Second, \ours can be combined with the other methods.
As our specificity metric is single-modality filtering and orthogonal to cross-modality filtering, such as CLIP filtering, all other baselines rely on, it can properly filter underspecified examples as shown in \Cref{fig:overview}.
This characteristic helps us to mark the first rank in the DataComp small and medium track by combining \ours with DFN.

\begin{wrapfigure}{r}{.53\linewidth}
    \vspace{-2em}
    \centering
    \includegraphics[width=\linewidth]{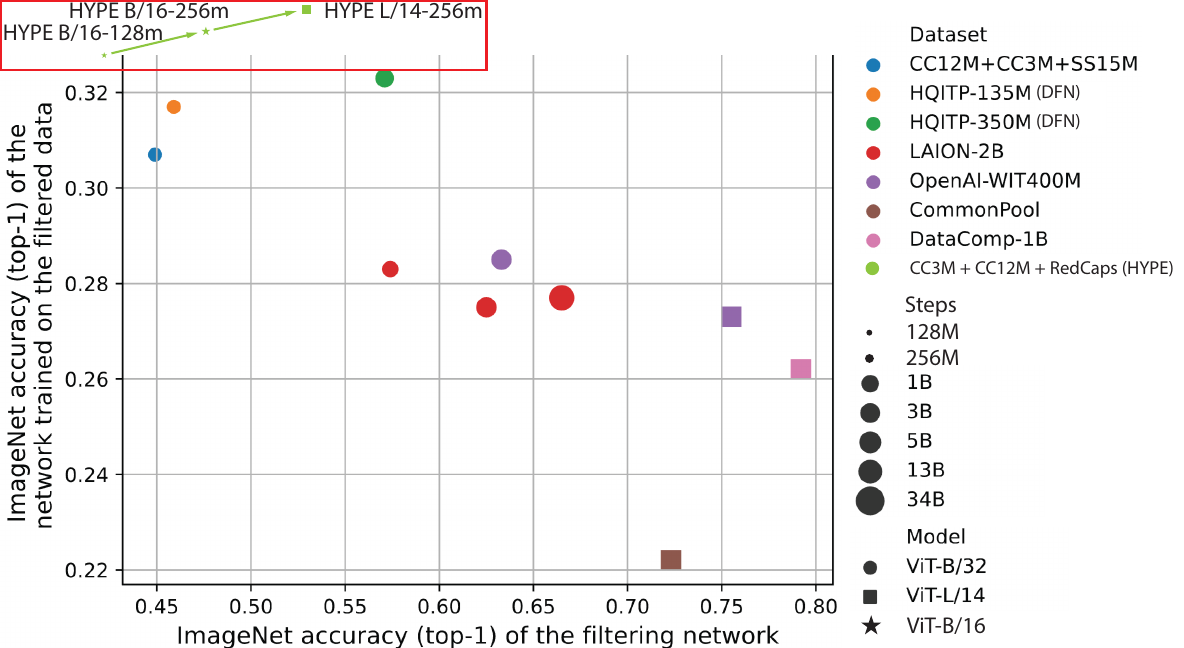}
    \caption{Filtering network IN-ZS acc. vs. Induced networks IN-ZS acc. (overlaid on the figure of the DFN paper \cite{dfn})}
    \label{fig:rebuttal_fig}
    \vspace{-2em}
\end{wrapfigure}

Additionally, we trained two more B/16 models with different training seen samples, 128M and 256M, whose IN-ZS accuracies are 42.3\% and 47.6\%.
With these models, we report their correlation to DataComp medium's IN-ZS accuracy as described in \cite{dfn}.
As shown in \cref{fig:rebuttal_fig}, a better-performing MERU model consistently induces a more effective filtering network, evidenced by improved zero-shot accuracy.
This upward trend suggests that further improvements in MERU could lead to even more effective filtering, which is not observed in the downward trend of Euclidean CLIP.

\subsection{Image-Only Contrastive Pre-training}
As our specificity metric is an uni-modal metric, unlike the CLIP similarity, we can apply our filtering method to uni-modal datasets.
Specifically, we investigate the image specificity metric (\enti)-based filtering for image-only self-supervised learning (SSL) methods, as previous works highlight the significance of iconic images in SSL training \cite{demystifying}.
Since \enti can efficiently identify iconic images (as shown in \Cref{fig:entailment_and_specificity}), we can expect that \enti-based filtering will lead to better SSL performances. 
We filter out underspecified images from the DataComp medium dataset and measure the SSL performances using two methods, SimCLR \cite{simclr}, and MoCo-v3 \cite{mocov3}.
We also provide CLIP similarity \cossim-based filtering, recognized for inducing well-aligned images from noisy datasets, based on image-caption pairs in the DataComp medium dataset.

\begin{table}[t]
\setlength{\tabcolsep}{15pt}
\centering
\caption{
ImageNet-1K linear probing classification accuracies of ViT-S. The table compares \cossim and \enti on inducing various size image-only datasets (DataComp Medium) for image-only self-supervised learning methods: SimCLR \cite{simclr} and MoCo-v3 \cite{mocov3}.
}
\vspace{-1em}
\begin{adjustbox}{width=\textwidth,center}
\begin{tabular}{@{}llcccc@{}}
\toprule
\multirow{2}{*}{Model} & \multirow{2}{*}{Filtering Metric} & \multicolumn{4}{c}{Dataset Size} \\
& & 0.13M & 0.32M & 0.64M & 1.28M \\ \midrule
\multirow{2}{*}{SimCLR \cite{simclr}}   & \cossim       & 47.06 & 45.47 & 52.31 & 49.68 \\
                          & \enti        & \textbf{53.30} & \textbf{50.89} & \textbf{57.49} & \textbf{54.55} \\
\multirow{2}{*}{MoCo-v3 \cite{mocov3}}  & \cossim       & 39.00 & 45.00 & 51.10 & 53.40 \\
                          & \enti        & \textbf{44.70} & \textbf{51.60} & \textbf{56.80} & \textbf{59.70} \\
\bottomrule
\end{tabular}
\end{adjustbox}
\label{tab:image_only}
\end{table}

\Cref{tab:image_only} shows the results from 1.28M images (comparable to ImageNet \cite{imagenet}) to 0.13M (10\% of ImageNet).
For the comparison, we use the established hyperparameters searched on ImageNet.
Following the practice of the DataComp filtering track, we keep the number of seen samples fixed for every dataset size, \ie, we use more epochs for smaller dataset sizes.
We report the linear probing performances on the ImageNet validation set following the standard SSL evaluation protocol.
\Cref{tab:image_only} reveals that \enti consistently outperforms \cossim across all dataset sizes and models.
Note that MoCo-v3 trained with a dataset induced by \enti outperforms SimCLR trained with a dataset induced by \cossim for the most of dataset sizes.
This result shows that the lower-performing SSL method can outperform the higher-performing ones by simply replacing the data.
\section{Discussion and Future Work}
We conclude this paper by discussing the limitations of our method and outlining future research directions. A notable limitation is that our experiments did not include the larger DataComp subsets, specifically the large and xlarge scales. Considering that \ours shows an increasing performance gap as the dataset size grows—from small to medium—it is reasonable to hypothesize that \ours might demonstrate exceptional performance when applied to these larger datasets.

Furthermore, \ours was designed with a hyperbolic CLIP size set to L/14, aligning with Datacomp's standards. However, there is a strong basis to believe that employing a larger hyperbolic CLIP architecture could significantly enhance performance metrics. Additionally, our research solely utilized \enti to create an image-only dataset. We posit that employing \entt to generate a text dataset could result in a visually meaningful text corpus. This new corpus could be instrumental in training a language model capable of rapidly adapting to visual inputs. Finally, we recognize the potential for extensive ablation studies, especially regarding the coefficient used in merging metrics for \ours, such in-depth analysis could yield further insights into the filtering model's behavior and performance, thereby enhancing its overall efficacy.

{
    \small
    \renewcommand{\bibname}{\protect\leftline{References}}
    \renewcommand\bibpreamble{\vspace{-4\baselineskip}}
    \setlength{\bibsep}{0pt}
    \bibliographystyle{ieeenat_fullname}
    \bibliography{main}
}

\appendix
\numberwithin{equation}{section}
\numberwithin{figure}{section}
\numberwithin{table}{section}
\section*{Appendix}
\section{Histograms}
\label{sec:histograms}
\begin{center}[t]
    \centering
    \captionsetup{type=figure}
    \vspace{-1em}
    \includegraphics[width=\linewidth]{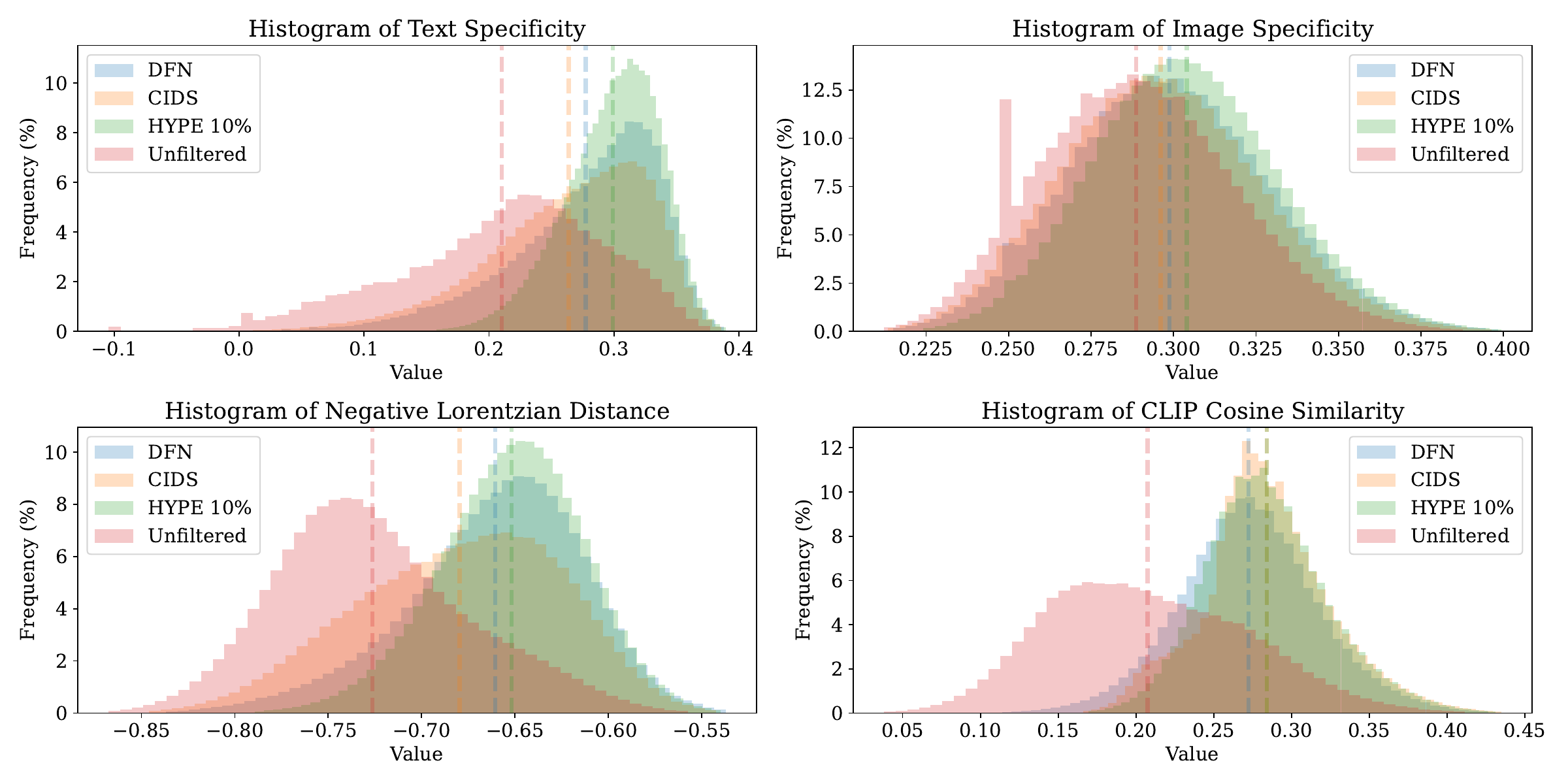}
    \captionof{figure}{In this paper, we examine four key metrics: text specificity (\entt), image specificity (\enti), negative Lorentzian distance (\negl), and CLIP cosine similarity (\cossim). For each metric, we present histograms to illustrate their distribution across various subsets of the Datacomp medium data pool. These subsets are color-coded for clarity: DFN \cite{dfn} is shown in blue, CIDS \cite{devil} in orange, our method (\ours) in green, and the no-filter condition in red. Additionally, we highlight their average values for each metric with vertical dotted lines in the respective histograms. Each method results in a different amount of data in the subset, so for ease of comparison, the y-axis shows the relative percentage rather than the count.}
    \label{fig:histograms}
\end{center}

In \Cref{fig:histograms}, we employed histograms to visually examine the alignment of each filtering method's subset with the metrics investigated in our study: text specificity (\entt), image specificity (\enti), negative Lorentzian distance (\negl), and CLIP cosine similarity (\cossim).
For all metrics except CLIP cosine similarity, a distinct hierarchy emerged among the methods: Unfiltered was the least aligned, followed by CIDS, then DFN, and HYPE showing the highest alignment.
The histogram results closely reflect the uniform sampling approach adopted for the subsets, as detailed in the \emph{uniform} column.
Uniform sampling ensures each sample in the subset is selected with equal probability.
However, when the sampling method strays from this uniform approach, such as the duplication sampling in the CIDS \cite{devil}, the histograms will lean towards data points that are sampled more frequently.
Such deviations could significantly impact the representativeness of the histograms.

This ranking aligns well with our expectations, given that \ours method explicitly filters the dataset based on these metrics.
However, the histogram of DFN \cite{dfn} is notably intriguing.
Despite the training of its filtering network focused solely on high-quality image-text pairs through a contrastive approach, DFN demonstrates considerable alignment with text and image specificity, as well as hyperbolic similarity.
This suggests that high-quality datasets might inherently possess high specificity, which inadvertently aligns with the objectives of our filtering approach that values high image and text specificity.

\section{More Qualitative Results}
\label{sec:more_quals}

In the remaining sections of the supplementary material, additional examples corresponding to the specificities are presented.
To enhance understanding of the relationship between \cossim, \entt, and \enti, we arranged the samples to be ordered from the left to the right, progressing from the lowest to the highest \cossim values.
This layout aids in visualizing the concept mentioned in the paper, demonstrating that with increasing \cossim values, image samples tend to contain texts, aligning more closely with those used in the OCR tasks.

\begin{figure*}[t]
    \centering
    \includegraphics[width=0.8\linewidth]{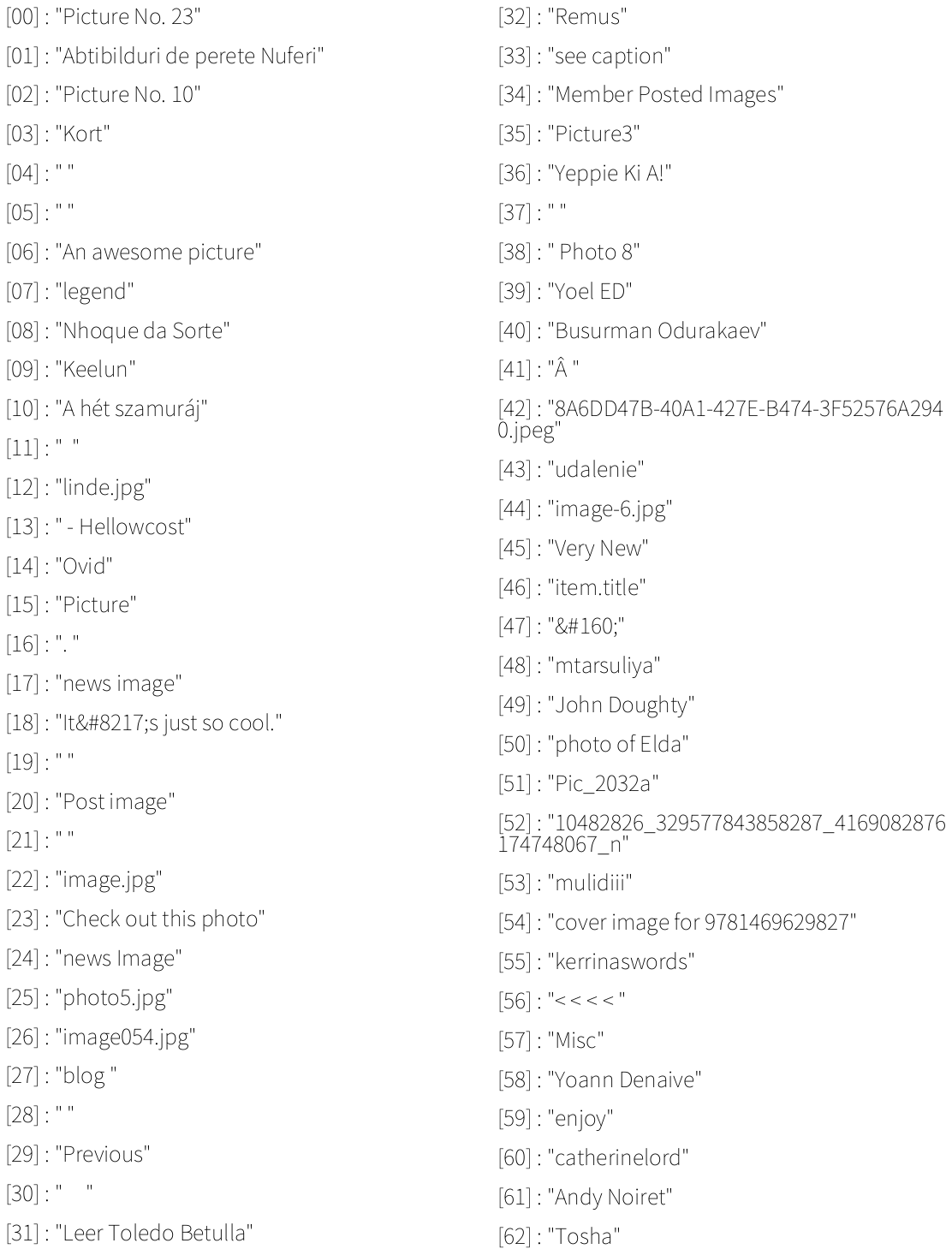}
    \caption{More example texts from Datacomp small that fall within the percentile of $p(\epsilon_t) < 1\%$. To ensure uniqueness, we removed duplicates, so each text, despite appearing as empty space in the figures, contains distinct characters. The examples are sequentially sampled in ascending order of CLIP scores within the given percentile pool.}
\end{figure*}

\begin{figure*}[t]
    \centering
    \includegraphics[width=0.8\linewidth]{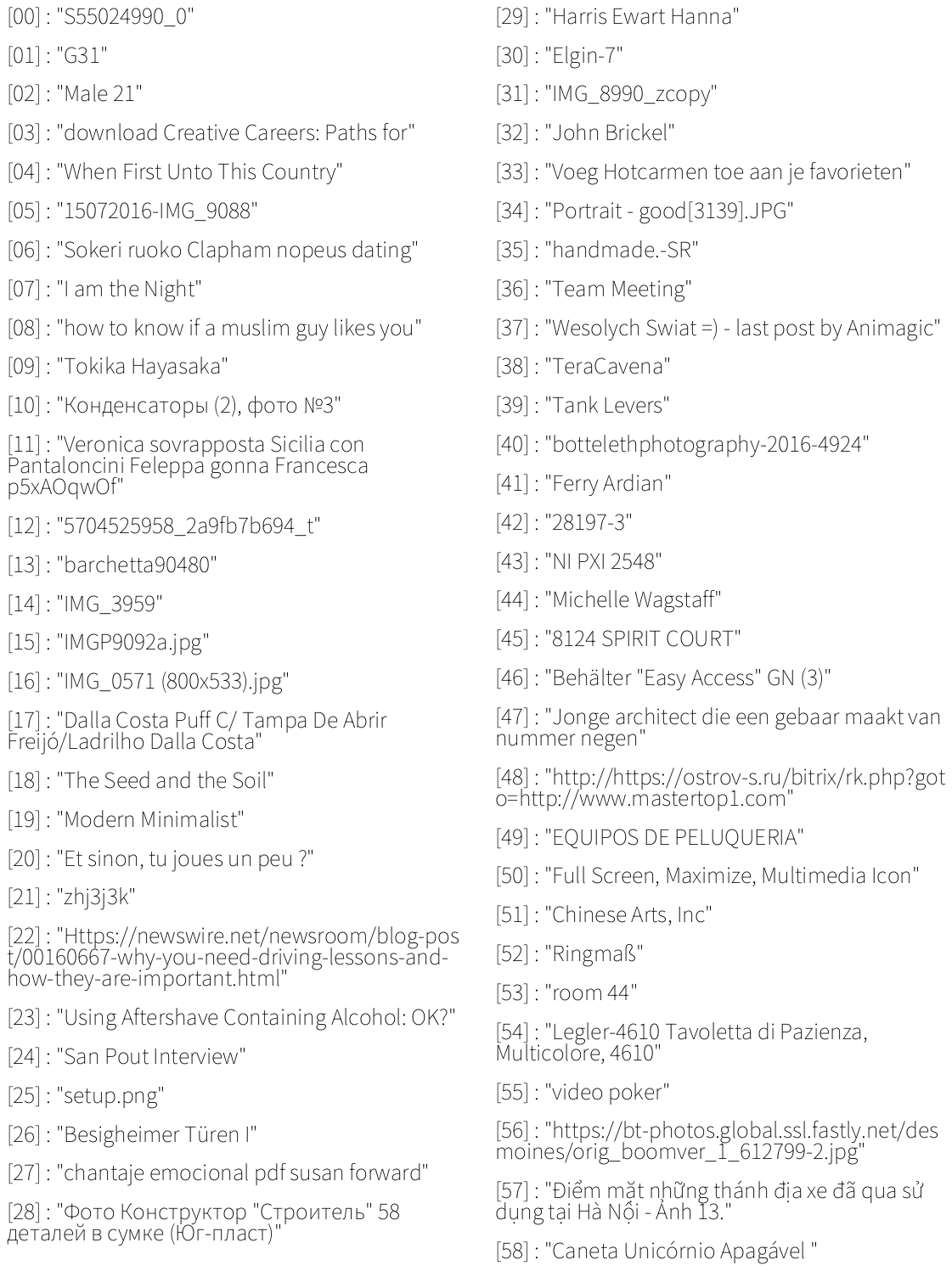}
    \caption{More example texts from Datacomp small that fall within the percentile of $19\% < p(\epsilon_t) < 21\%$. The examples are sequentially sampled in ascending order of CLIP scores within the given percentile pool.}
\end{figure*}

\begin{figure*}[t]
    \centering
    \includegraphics[width=0.8\linewidth]{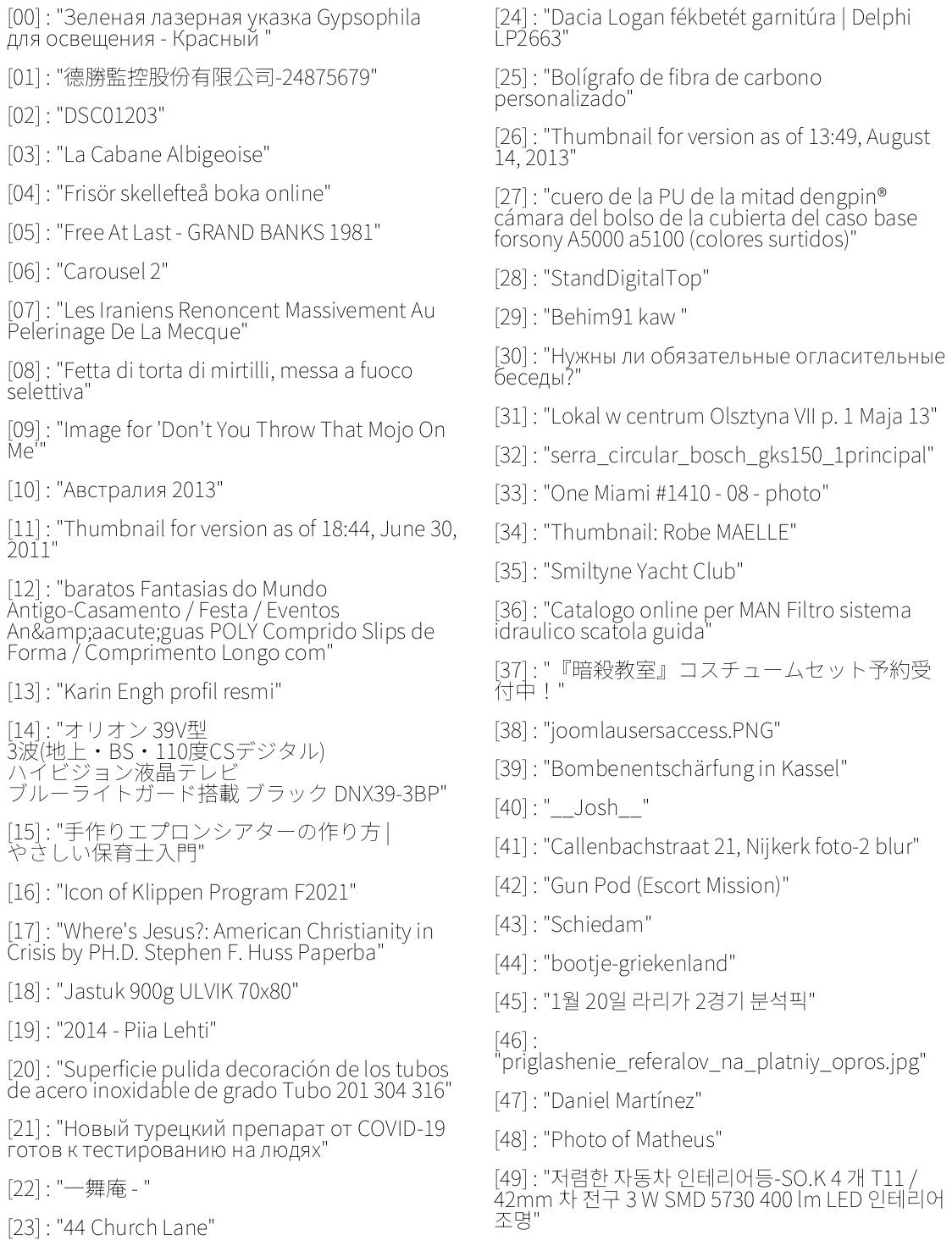}
    \caption{More example texts from Datacomp small that fall within the percentile of $39\% < p(\epsilon_t) < 41\%$. The examples are sequentially sampled in ascending order of CLIP scores within the given percentile pool.}
\end{figure*}

\begin{figure*}[t]
    \centering
    \includegraphics[width=0.8\linewidth]{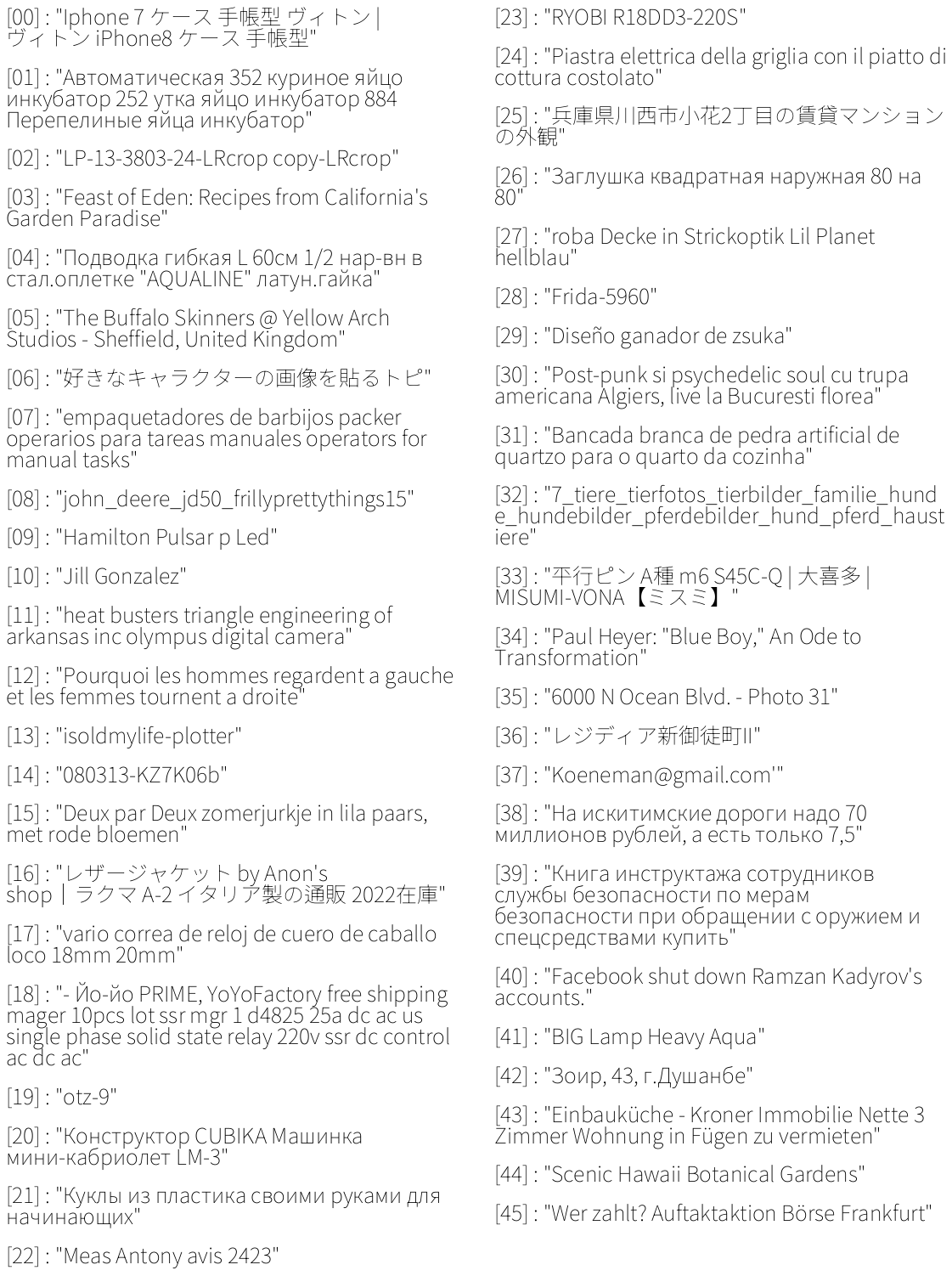}
    \caption{More example texts from Datacomp small that fall within the percentile of $59\% < p(\epsilon_t) < 61\%$. The examples are sequentially sampled in ascending order of CLIP scores within the given percentile pool.}
\end{figure*}

\begin{figure*}[t]
    \centering
    \includegraphics[width=0.8\linewidth]{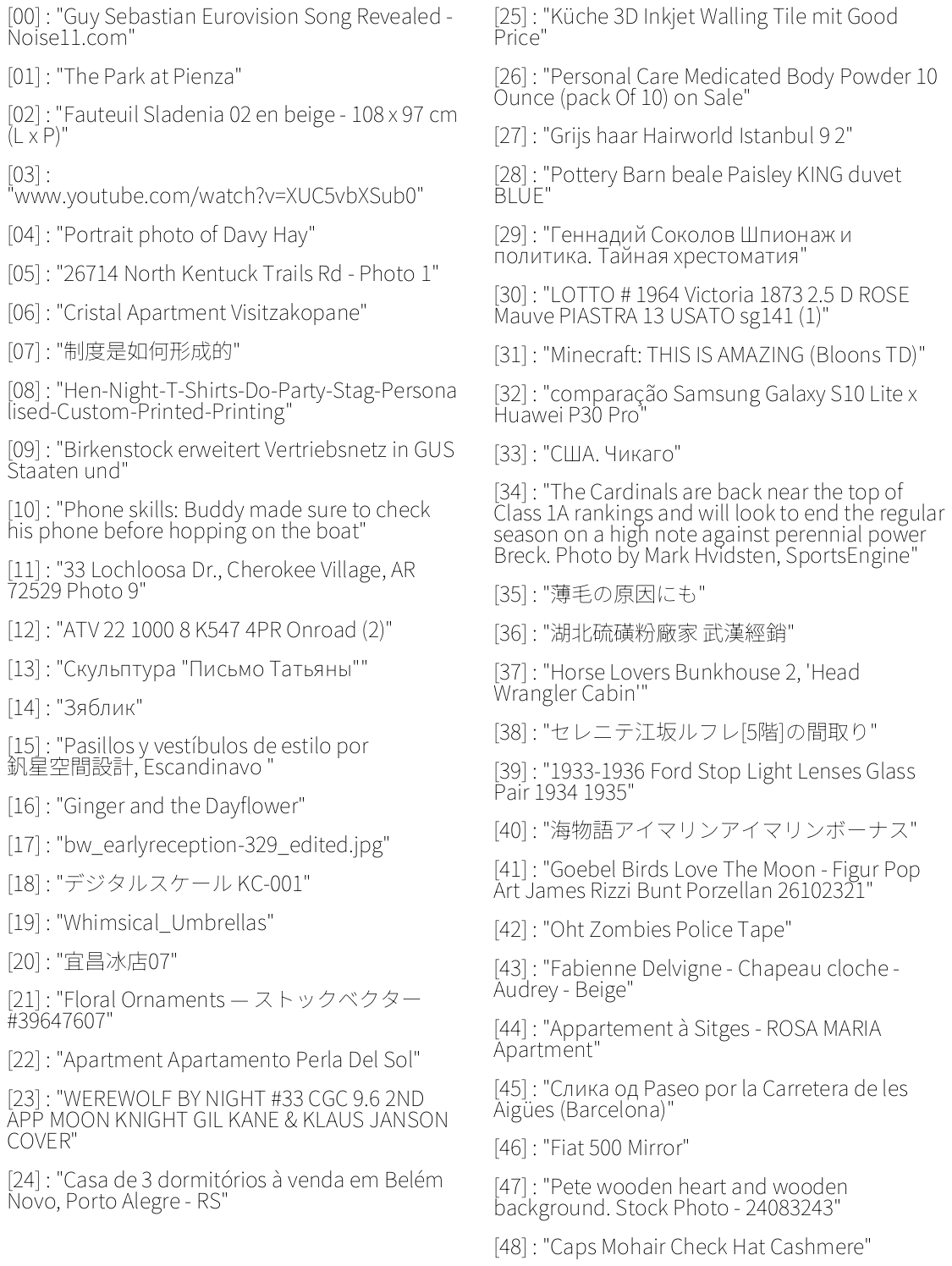}
    \caption{More example texts from Datacomp small that fall within the percentile of $79\% < p(\epsilon_t) < 81\%$. The examples are sequentially sampled in ascending order of CLIP scores within the given percentile pool.}
\end{figure*}

\begin{figure*}[t]
    \centering
    \includegraphics[width=0.8\linewidth]{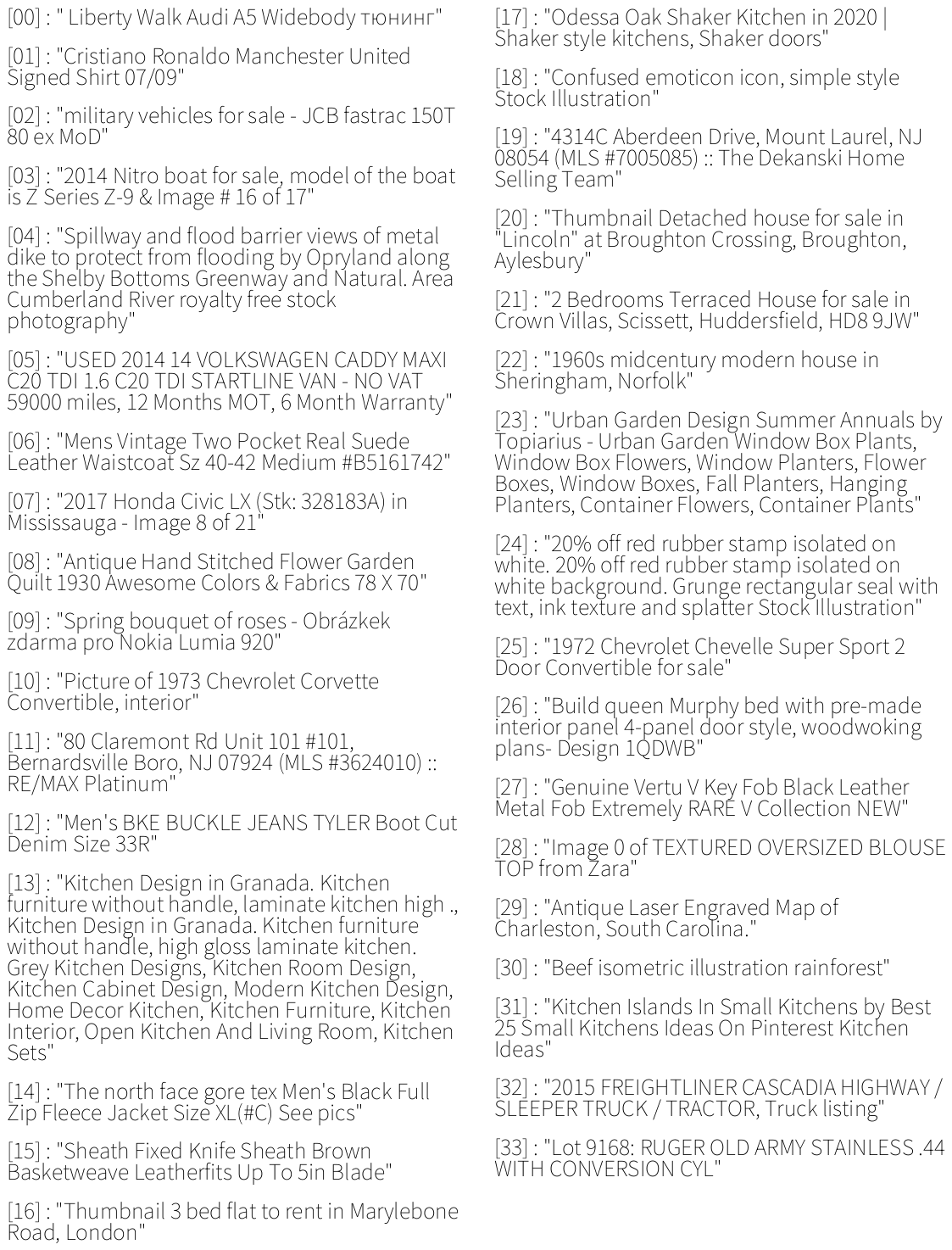}
    \caption{More example texts from Datacomp small that fall within the percentile of $99\% < p(\epsilon_t)$. The examples are sequentially sampled in ascending order of CLIP scores within the given percentile pool.}
\end{figure*}

\begin{figure*}[t]
    \centering
    \includegraphics[width=0.8\linewidth]{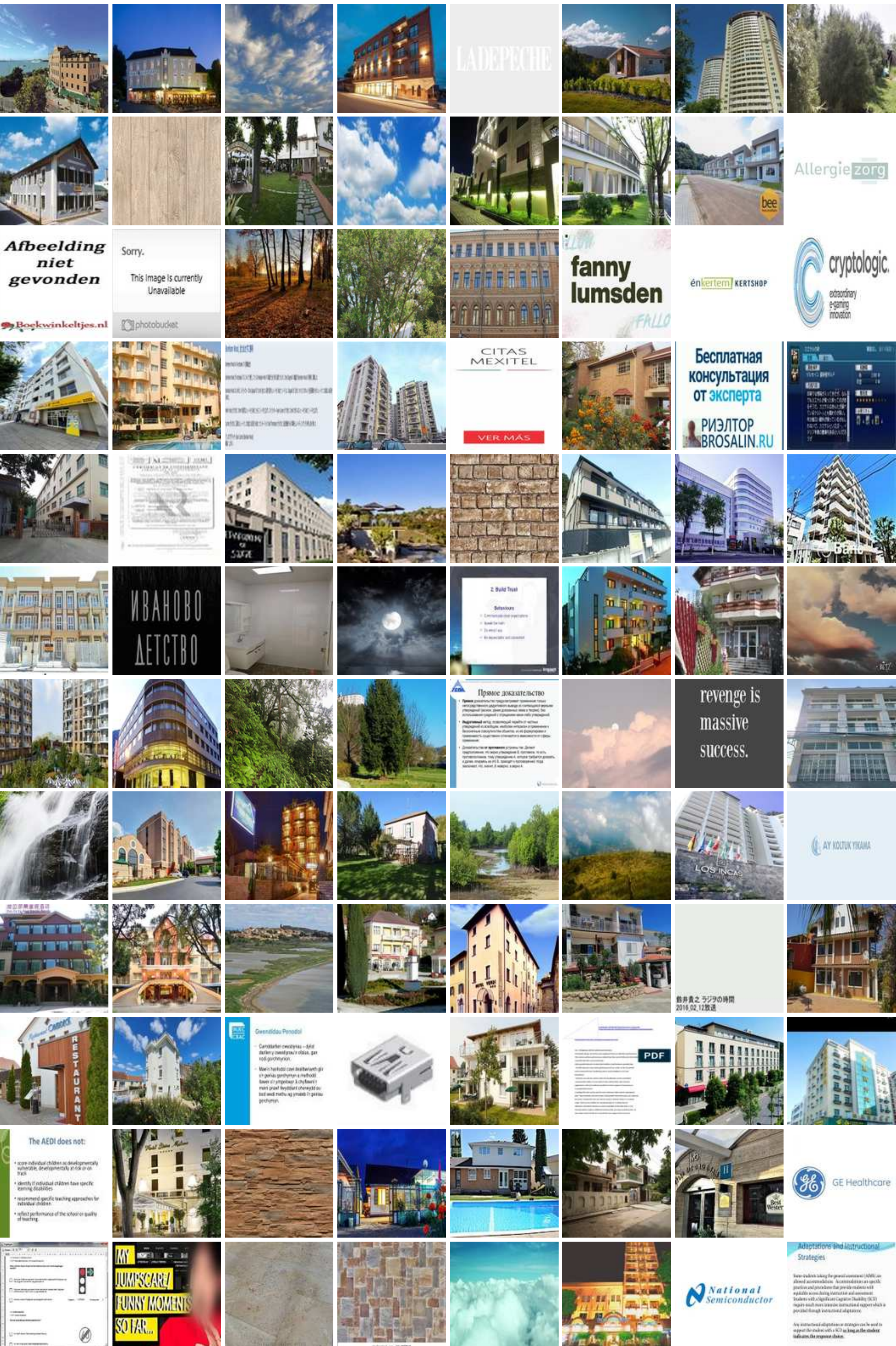}
    \caption{More example images from Datacomp small that fall within the percentile of $p(\epsilon_i) < 1\%$. The examples are sequentially sampled in ascending order of CLIP scores within the given percentile pool.}
\end{figure*}

\begin{figure*}[t]
    \centering
    \includegraphics[width=0.8\linewidth]{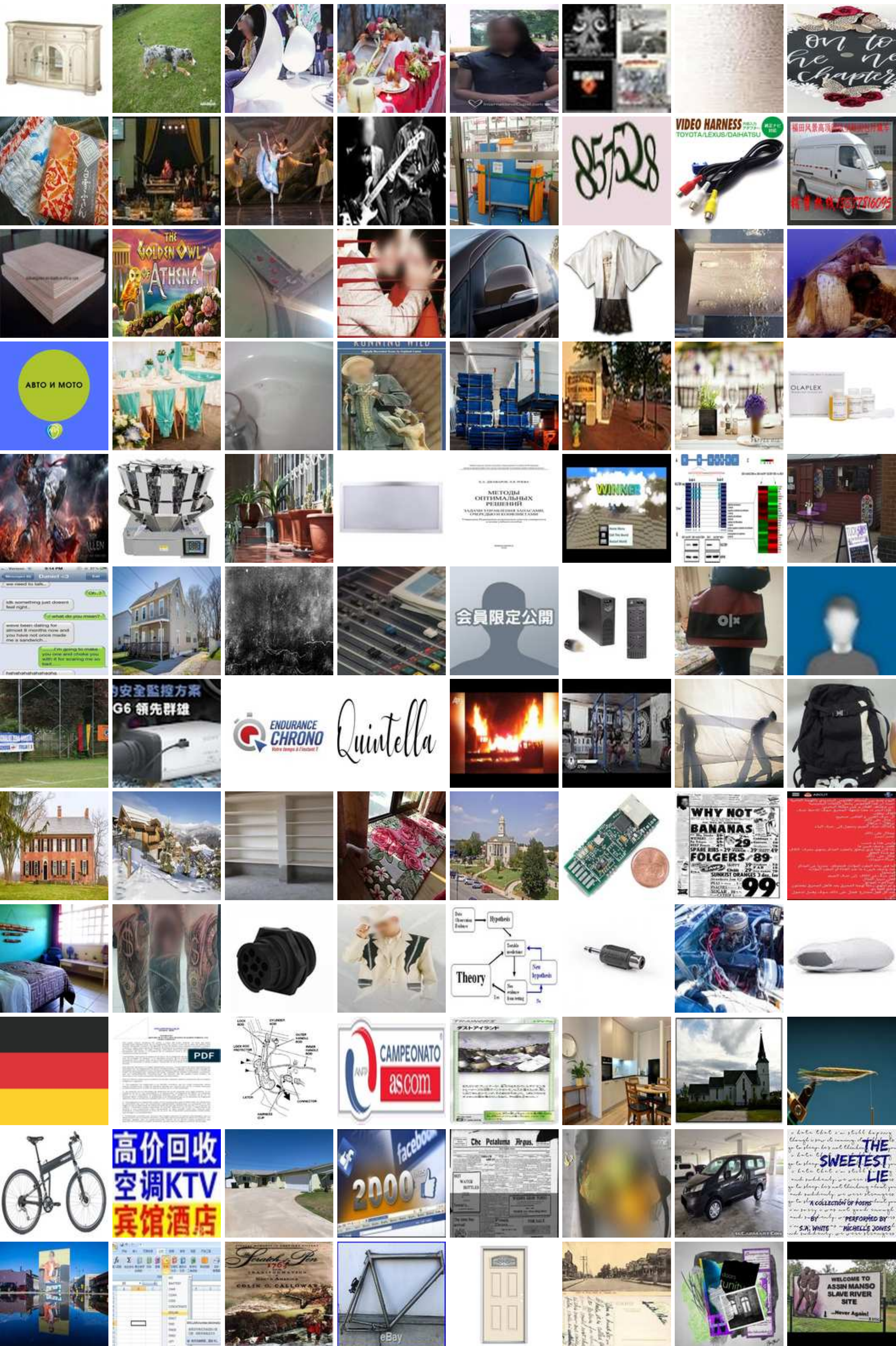}
    \caption{More example images from Datacomp small that fall within the percentile of $19\% < p(\epsilon_i) < 21\%$. The examples are sequentially sampled in ascending order of CLIP scores within the given percentile pool.}
\end{figure*}

\begin{figure*}[t]
    \centering
    \includegraphics[width=0.8\linewidth]{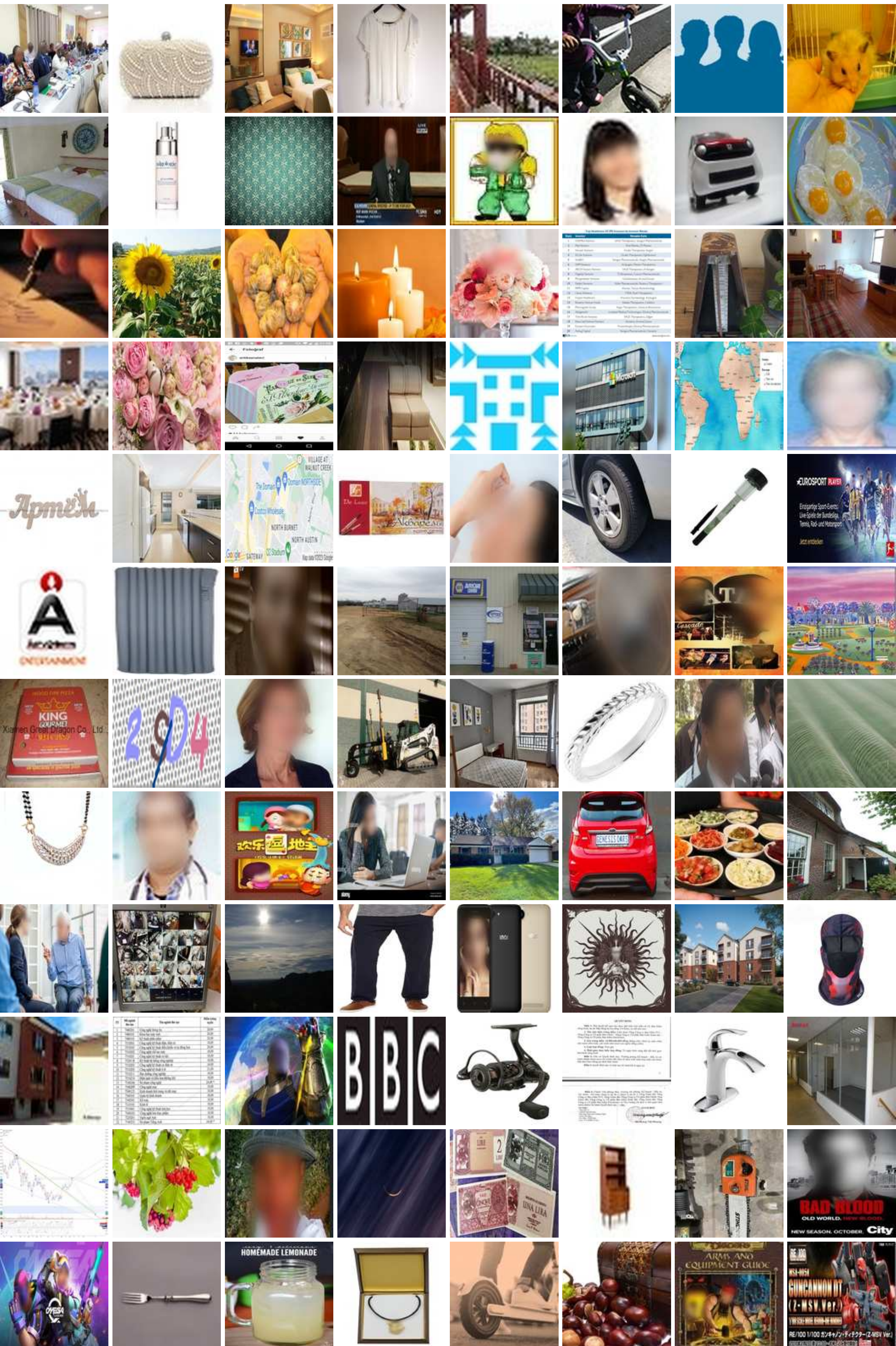}
    \caption{More example images from Datacomp small that fall within the percentile of $39\% < p(\epsilon_i) < 41\%$. The examples are sequentially sampled in ascending order of CLIP scores within the given percentile pool.}
\end{figure*}

\begin{figure*}[t]
    \centering
    \includegraphics[width=0.8\linewidth]{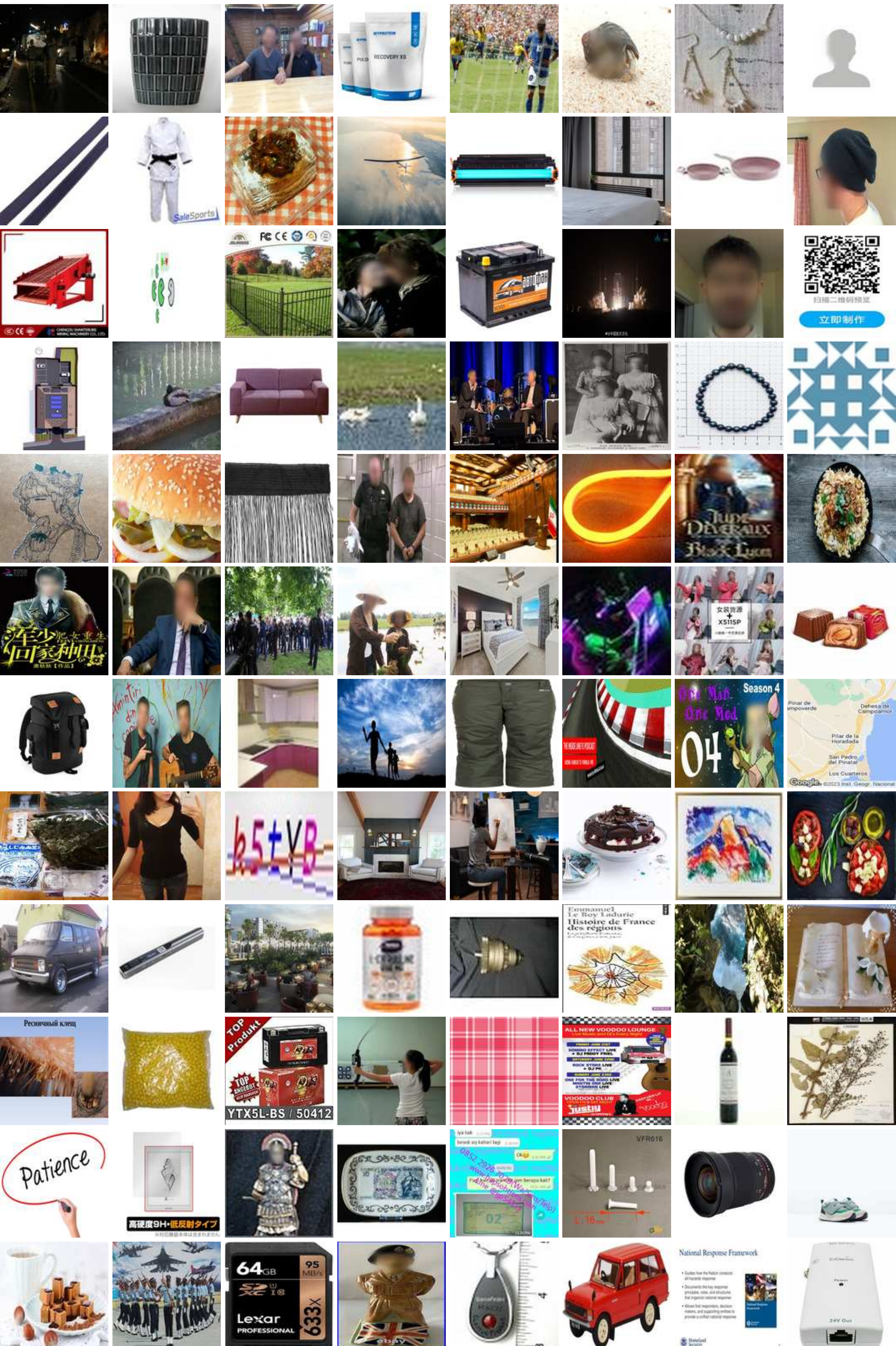}
    \caption{More example images from Datacomp small that fall within the percentile of $59\% < p(\epsilon_i) < 61\%$. The examples are sequentially sampled in ascending order of CLIP scores within the given percentile pool.}
\end{figure*}

\begin{figure*}[t]
    \centering
    \includegraphics[width=0.8\linewidth]{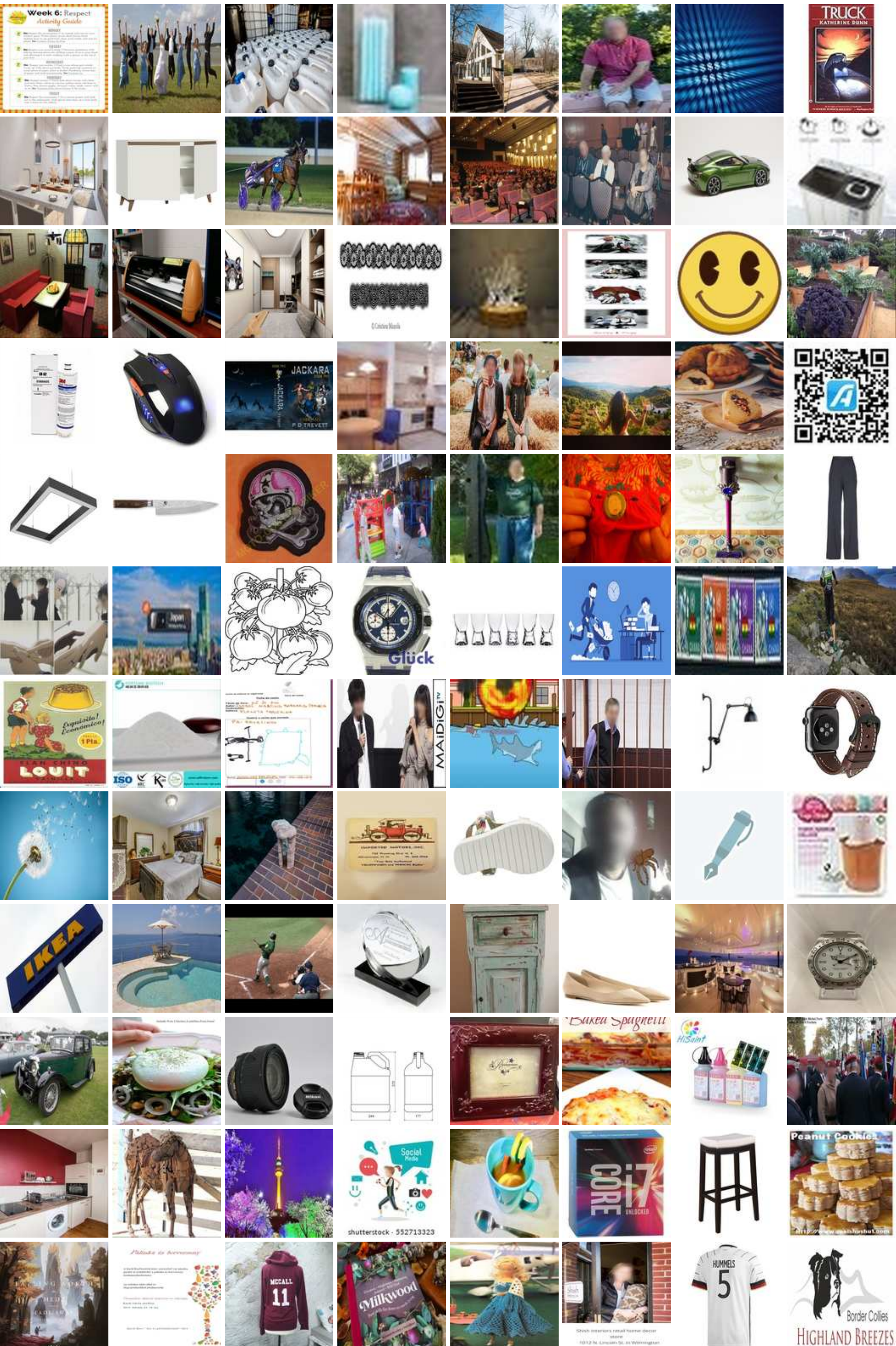}
    \caption{More example images from Datacomp small that fall within the percentile of $79\% < p(\epsilon_i) < 81\%$. The examples are sequentially sampled in ascending order of CLIP scores within the given percentile pool.}
\end{figure*}

\begin{figure*}[t]
    \centering
    \includegraphics[width=0.8\linewidth]{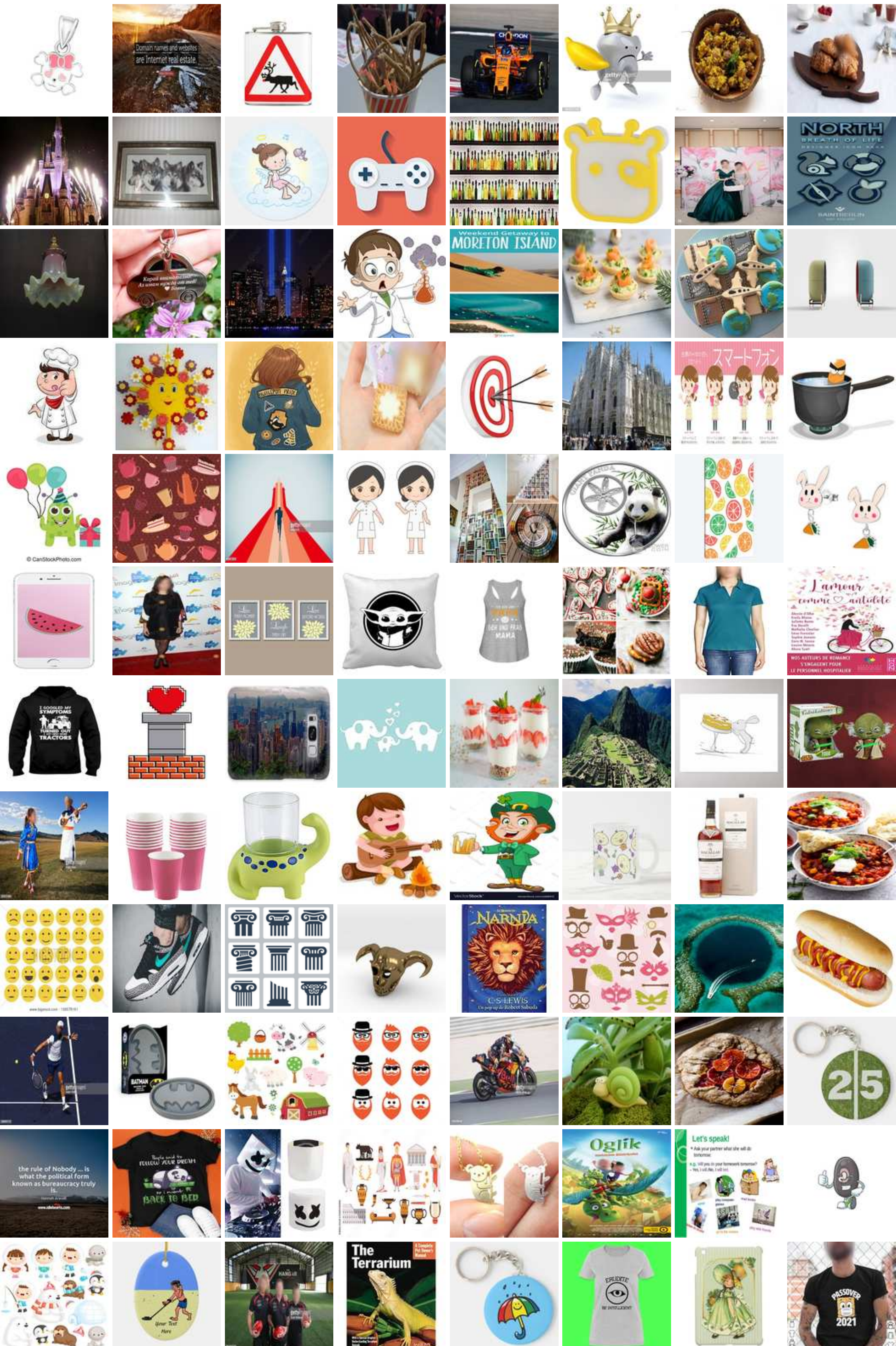}
    \caption{More example images from Datacomp small that fall within the percentile of $99\% < p(\epsilon_i)$. The examples are sequentially sampled in ascending order of CLIP scores within the given percentile pool.}
\end{figure*}
\end{document}